\documentclass[runningheads]{llncs}
\usepackage{graphicx}
\usepackage{comment}
\usepackage{amsmath,amssymb} 
\usepackage{color}
\usepackage{kotex}

\usepackage{xcolor, colortbl}
\usepackage{mathtools}
\usepackage{tabularx}
\usepackage{multirow}
\usepackage{enumitem}
\usepackage{bbm}
\usepackage{wrapfig}
\usepackage{booktabs}
\usepackage[pagebackref=true,breaklinks=true,letterpaper=true,colorlinks,bookmarks=false]{hyperref}

\newcolumntype{C}[1]{>{\centering\let\newline\\\arraybackslash\hspace{0pt}}p{#1}}

\def\ie{\emph{i.e.}}
\def\eg{\emph{e.g.}}
\def\etal{\emph{et al.}}

\definecolor{brown}{rgb}{0.65, 0.16, 0.16}
\definecolor{purp}{rgb}{0.65, 0.16, 0.65}
\definecolor{orange}{rgb}{1.0, 0.5, 0.0}
\definecolor{blue}{rgb}{0.0, 0.5, 1.0}
\definecolor{green}{rgb}{0, 0.8, 0}
\definecolor{lgreen}{rgb}{0.6, 0.8, 0}
\definecolor{red}{rgb}{0.8, 0, 0}
\definecolor{darkblue}{rgb}{0, 0.2, 0.6}

\newcommand{\juwon}[1]{{\color{blue}{(#1)}}}

\newcommand{\green}[1]{{\color{green}{#1}}}
\newcommand{\red}[1]{{\color{red}{#1}}}
\newcommand{\sunghyun}[1]{{\color{purple}{sunghyun:(#1)}}}

\newcommand{\Fig}[1]{Fig.~\ref{fig:#1}}
\newcommand{\Sec}[1]{Sec.~\ref{sec:#1}}

\begin{document}
\pagestyle{headings}
\mainmatter
\def\ECCVSubNumber{960}  

\title{URIE: Universal Image Enhancement\\for Visual Recognition in the Wild}

\renewcommand{\thefootnote}{\fnsymbol{footnote}}
\author{
Taeyoung Son$^{1}$\hspace{0.2cm}
Juwon Kang$^{1}$\hspace{0.2cm}
Namyup Kim$^{1}$\hspace{0.2cm}
Sunghyun Cho$^{2}$\hspace{0.2cm}
Suha Kwak$^{2}$
}
\institute{
$^{1}$Department of Computer Science and Engineering\\
$^{2}$Graduate School of Artificial Intelligence\\
POSTECH, Pohang, Korea\\
{\tt\small \url{http://cvlab.postech.ac.kr/research/URIE/}}
}
\authorrunning{Taeyoung Son, Juwon Kang, Namyup Kim, Sunghyun Cho and Suha Kwak}

\maketitle

\begin{abstract}
Despite the great advances in visual recognition, it has been witnessed that recognition models trained on clean images of common datasets are not robust against distorted images in the real world.
To tackle this issue, we present a Universal and Recognition-friendly Image Enhancement network, dubbed URIE, which is attached in front of existing recognition models and enhances distorted input to improve their performance without retraining them.
URIE is universal in that it aims to handle various factors of image degradation and to be incorporated with any arbitrary recognition models.
Also, it is recognition-friendly since it is optimized to improve the robustness of following recognition models, instead of perceptual quality of output image. 
Our experiments demonstrate that URIE can handle various and latent image distortions and improve the performance of existing models for five diverse recognition tasks where input images are degraded. 
\end{abstract}

\section{Introduction}\label{sec:intro}



The development of deep learning has made the great advances in visual recognition. 
Especially, the recent advances in this field have been concentrated on accurate and comprehensive understanding of high-quality images taken in limited environments.
However, images in the real world are often corrupted by various factors including adverse weather conditions, sensor noise, under- or over-exposure, motion blur, and compression artifact.
Since most of existing recognition models do not consider these factors, their performance could be easily degraded when input image is corrupted~\cite{dirtypixel,Hendrycks2019_ImageNetC,Zendel_2018_ECCV}, which could be fatal in safety-critical applications like autonomous driving. 

A straightforward way to resolve this issue is to improve the quality of input image fed to recognition models by a preprocessing method.
Image restoration seems suited to this purpose at first glance, but unfortunately, it is not in general for the following reasons. 
First, most of restoration methods assume that images are degraded by a single and known factor of corruption.
They are thus hard to handle realistic recognition scenarios where images can be corrupted by various factors whose types and degrees are hidden. 
Second, they are not trained for visual recognition but for human perception.
Hence input images restored by them do not always guarantee performance improvement of recognition models as demonstrated in~\cite{dirtypixel,Li_2019_CVPR,Pei_2018_ECCV,vidal2018ug2}.
Finally, recent restoration networks tend to be heavy in computation, thus could make the entire system overly expensive when integrated with recognition models.

This paper addresses the problem through a novel image enhancement model dedicated to robust visual recognition in the wild. 
We call our model Universal and Recognition-friendly Image Enhancement network, dubbed URIE, since we keep the following properties in our mind when developing the model.
(1)~\emph{Universally applicable}: 
Our model aims to be attached in front of any existing recognition models to improve their robustness in any situations without retraining them. 
To this end, it has to deal with various factors of degradation whose types and intensities are latent, and be applicable to diverse recognition tasks and models. 
(2)~\emph{Recognition friendly}: 
Our model should be optimized for improving the performance of following recognition models, instead of making images looking plausible.
(3)~\emph{Computationally efficient}:
The complexity of our model has to be low to minimize the computational burden it additionally imposes.

\begin{figure*} [!t]
\centering
\includegraphics[width = 1 \textwidth]{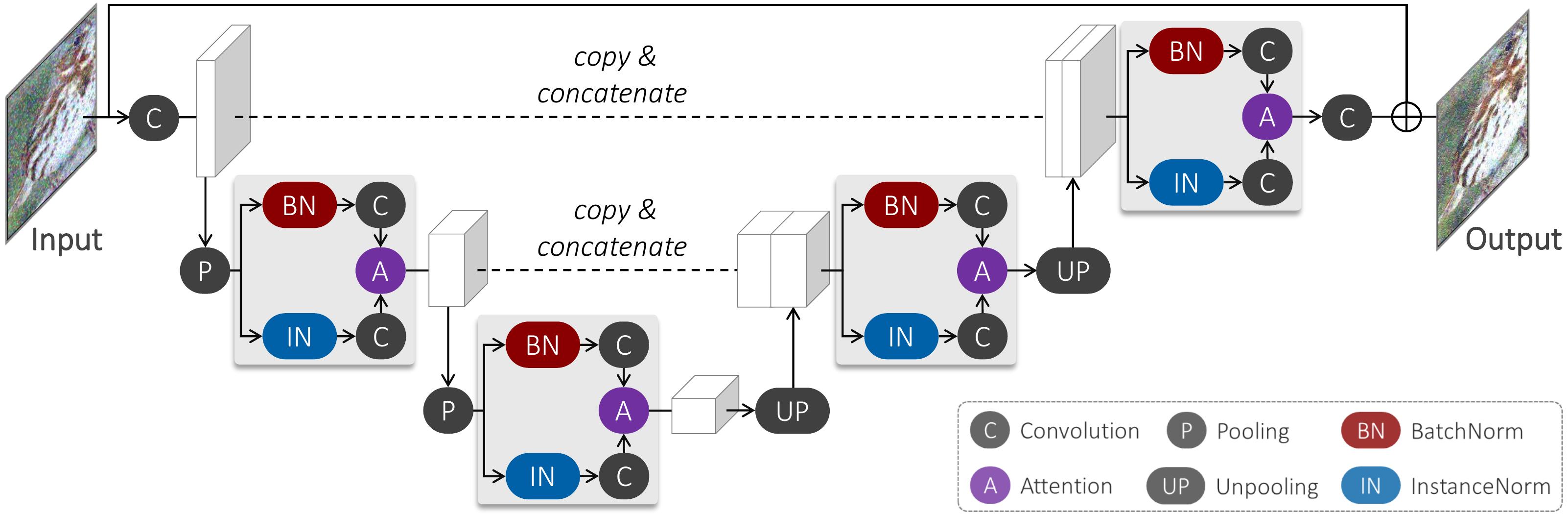}
\caption{
Overall architecture of URIE. The Selective Enhancement Modules (SEM) are indicated by gray rectangles. Details of these modules are illustrated in Fig.~\ref{fig:selective_path}.
} 
\label{fig:overall_arch}
\end{figure*}

We come up with a new network architecture and its own training strategy to implement these properties.
In particular, the architecture of URIE is devised to deal with various and latent distortions efficiently and effectively. 
To handle diverse image distortions, URIE runs a large number of different image enhancement procedures internally by a single feed-forward path.
At the same time, it has the capability to select enhancement procedures appropriate for dealing with the latent type of input distortion.
These features are given by Selective Enhancement Module (SEM), a basic building block of URIE.
The architecture of URIE composed of SEMs is illustrated in Fig.~\ref{fig:overall_arch}.
Each SEM provides two different steps towards image enhancement.
Hence, by concatenating SEMs, the network will have an exponentially large number of enhancement procedures.
Moreover, each SEM has an attention mechanism that assigns a larger attention weights to its enhancement step more appropriate to deal with the input distortion than the other; this enables URIE to dynamically select useful enhancement procedures depending on the latent type of distortion.

In addition, our new training method enables URIE to be recognition-friendly as well as universal.
During training, URIE is coupled to an image classification network pretrained on the original ImageNet dataset~\cite{Imagenet}.
It is then trained in an end-to-end manner to improve the performance of the classifier on distorted images.
As training data, we adopt ImageNet-C~\cite{Hendrycks2019_ImageNetC}, in which images of the original ImageNet are degraded by 19 classes of distortion.
This training strategy gives URIE the recognition-friendly property since the training objective is not the perceptual quality of output image but the recognition performance of the coupled classifier. 
In addition, we expect that training on such a large-scale dataset with a variety of corruptions allows our model to be less sensitive to the task and architecture of the following recognition model, analogously to ImageNet pretrained networks that have been used for various purposes~\cite{deconvnet,Oquab2014,vggnet} other than their original target.



The efficacy of URIE is evaluated on five different recognition tasks, in which recognition models are pretrained on clean images and tested with URIE on corrupted ones with no finetuning. 
Experimental results demonstrate that URIE improves recognition performance in the presence of image distortions including those unseen during training, and regardless of target tasks and recognition networks coupled with it. 
URIE is also compared with the state-of-the-art image restoration model~\cite{OWAN_CVPR19}, and outperforms it in terms of both recognition accuracy and computational efficiency.
In summary, our contribution is four-fold: 
\begin{itemize}
    \item We introduce a new method towards robust visual recognition. Our method improves the robustness of existing recognition models without retraining them through a single and universal image enhancement module. 
    \item We propose a novel network architecture for image enhancement that can handle various and latent types of image distortions effectively and efficiently.
    \item We present a new training strategy that enables our enhancement model to be universally applicable and recognition-friendly.
    \item In our experiments, our model improves the recognition performance of existing models substantially in the presence of various image distortions regardless of tasks and models it is tested on. 
\end{itemize}

%

\section{Related Work} \label{sec:relatedwork}


\subsection{Fragility of Visual Recognition Models}
Fragility of deep neural networks for visual recognition has been studied to comprehend their robustness against challenging conditions that they often encounter in the real world.
Diamond~\etal~\cite{dirtypixel} demonstrated that image classification networks are sensitive to noise and blur.
Similarly, Pei~\etal~\cite{Pei_2018_ECCV} showed that haze can degrade the performance of classification networks. 
On the other hand, Zendel~\etal~\cite{Zendel_2018_ECCV} defined various factors of image degradation appearing frequently in autonomous driving scenarios, and analyzed 
the robustness of existing semantic segmentation networks to them.
Hendrycks and Dietterich~\cite{Hendrycks2019_ImageNetC} studied performance degradation of image classifiers by simulated image distortions.
These results indicate that existing recognition networks trained on clean images are fragile in the presence of image distortion.
Geirhos~\etal~\cite{Geirhos2018} showed that one way to resolve this issue is to finetune models to distorted images but such models are not well generalized when tested on other distortion types.

Our work is motivated by these observations. We address the above problem through an image enhancement model that deals with a variety of image distortions and is optimized to improve the robustness of recognition models.

\subsection{Recognition of Distorted Images}

Previous methods to robust recognition of distorted images can be grouped into two classes:
\emph{direct recognition} and \emph{recognition-friendly image enhancement}.

\noindent \textbf{Direct recognition of corrupted images.}
Classical methods in this category rely on visual features robust against image distortions like blur-invariant features~\cite{blur_face,blur_track}.
Recent methods focus on developing new network architectures and training strategies dedicated to the robust recognition.
Wang~\etal~\cite{Wang_VRR2016} and Singh~\etal~\cite{Singh_2019_ICCV} developed neural networks for classifying low resolution images,
and Lee~\etal~\cite{vpgnet} proposed a multi-task network for understanding road markings in rainy days and nighttime.
Also, domain adaptation techniques have been widely adopted for corruption-aware training of recognition models.
Considering different types of degradation as individual domains, they allow models for object detection and semantic segmentation to be adapted to adverse weather conditions~\cite{Chen_2018_CVPR,Sakaridis_2018_ECCV,Wu_2019_ICCV} and nighttime images~\cite{Sakaridis_2019_ICCV}.

\noindent \textbf{Recognition-friendly image enhancement.}
Models in this category are designed and trained to enhance input images so that it becomes more informative and task-optimized for following recognizers.
Diamond~\etal~\cite{dirtypixel} proposed a differentiable image processing module learned jointly with a classifier for improved classification of noisy and blurred images.
Similarly, Liu~\etal~\cite{denoising_meets_vision} developed denoising networks for classification and semantic segmentation of noisy images.
Gomez~\etal~\cite{Gomez_ICRA_2018} studied network architectures that enhance images taken in difficult illumination conditions for robust visual odometry.
Sharma~\etal~\cite{Sharma_2018_CVPR} investigated a way to dynamically predict image processing filters that are applied to input images and make them more classification-friendly.

The aforementioned methods of both categories are limited in terms of universality as they handle a single degradation type and/or are optimized for only one recognition task.
On the other hand, our goal is to build a single enhancement model that can cope with various types of image degradation and be attached to any architectures for any recognition tasks.


\subsection{Image Restoration}
Most of image restoration methods focus on a single type of image distortion 
(\eg, denoising~\cite{dncnn}, deblurring~\cite{nah2017deep}, dehazing~\cite{Li_2017_ICCV}, deraining~\cite{Yasarla_2019_CVPR}, and super-resolution~\cite{dong2016image}),
and consequently they are incapable of handling real world images where the type and intensity of distortion are hidden in general.
A few network architectures for image restoration are designed to handle various types of degradation~\cite{liu2018non,tai2017memnet}, but not at the same time as they have to be trained for each restoration objective. 
Recently, the above issue of conventional image restoration has been addressed by new architectures that can deal with various and latent factors of degradation~\cite{OWAN_CVPR19,RL-Restore_CVPR18}.
However, they are not optimized for image recognition, thus may not guarantee improved performance of coupled visual recognition models. 
In addition, since they are computationally heavy, they will increase the complexity of overall system significantly if integrated.


\section{URIE: Architecture and Training} \label{sec:method}

Our final target in this paper is to improve the robustness of existing recognition models to diverse image distortions \emph{without retraining them}.
URIE is proposed to address this challenging problem by converting an input image degraded by latent distortion into another that can be recognized more reliably by any existing models. 
To this end, it is designed and trained to be universally applicable to various image distortions and recognition architectures, while being recognition-friendly for improved performance of the following recognition models.

Specifically, URIE is devised to derive many different enhancement procedures from a single feed-forward architecture, and at the same time, to dynamically select procedures useful to handle the input distortion. 
Thanks to these features, URIE can be universally applicable to a large variety of image distortions whose types and intensities are latent.
This architecture is embodied by a series of basic building blocks, which we call SEMs.
Furthermore, we provide a new training method that enables URIE to be recognition-friendly and universally applicable to any recongnition models.

The remainder of this section presents details of SEM, discusses overall architecture and other design points of URIE, and describes our training strategy that makes URIE universally applicable and recognition-friendly.

\subsection{Selective Enhancement Module}
\label{sec:selective_enhc}

\begin{figure*} [!t]
\centering
\includegraphics[width = 1 \textwidth]{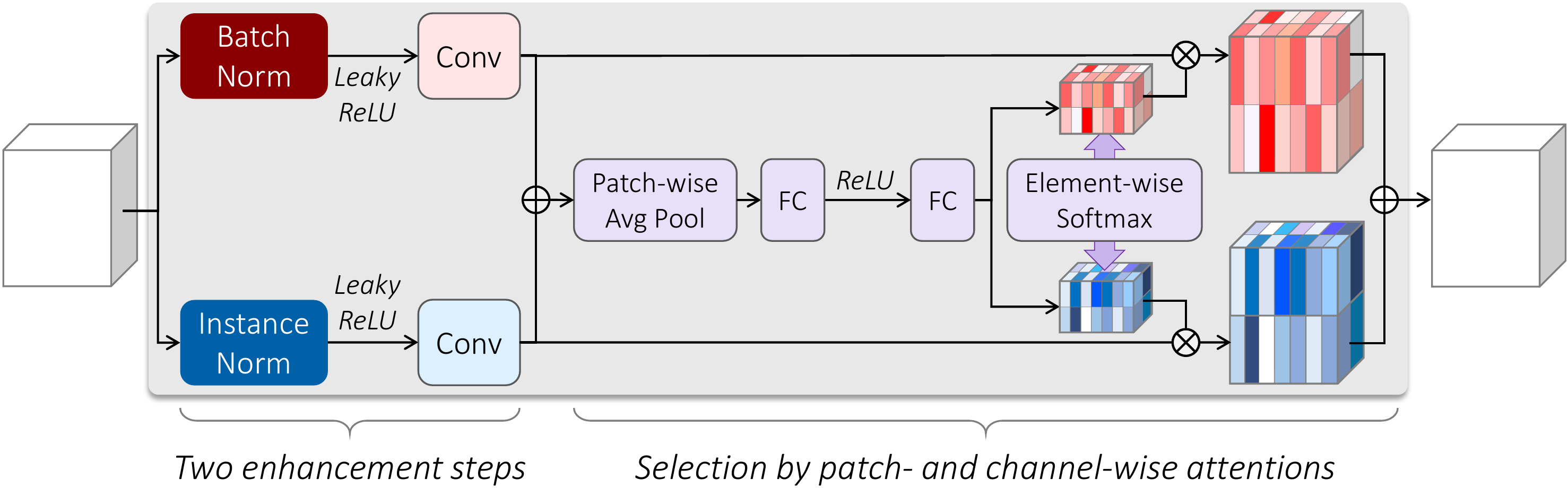}
\caption{
Details of SEM. $\oplus$ and $\otimes$ indicate element-wise summation and multiplication between feature maps, respectively.
} 
\label{fig:selective_path}
\end{figure*}

In URIE, an image enhancement procedure is represented as a sequence of convolution layers.
As a basic building block of the network, SEM provides individual steps towards image enhancement (\ie, convolution layers) so that a series of such steps forms an enhancement procedure.
To be specific, a single SEM presents two diverse enhancement steps, and has the capability to estimate which one among them is more appropriate to deal with the input distortion.
The detailed architecture of SEM is illustrated in Fig.~\ref{fig:selective_path}.

The two enhancement steps in SEM have separate convolution layers associated with different normalization operations, Instance Normalization (IN)~\cite{Instancenorm} and Batch Normalization (BN)~\cite{Batchnorm}.
We adopt these two different normalization operations to learn a pair of diverse and complementary enhancement functions. 
IN has been used in neural style transfer to normalize the style of image while keeping its major content by normalizing statistics per image independently.
Regarding a distortion type as an image style, we expect the enhancement step with IN to moderate the effect of distortion for individual images. 
On the other hand, the enhancement step with BN maintains the distortion information disregarded by the other, and is trained by considering a variety of distorted images together within each minibatch. 
We thus expect that it is complementary to its IN counterpart and useful for understanding input distortion.
Empirical justification of using IN and BN can be found in Sec.~\ref{sec:abliation_norm} of the appendix. 

The two outputs of these enhancement steps are fed to the attention module that determines which one among them is more useful to deal with the input distortion for each channel and each local patch of the input image.
Let $F_\textrm{in} \in \mathbb{R}^{w \times h \times C}$ and $F_\textrm{bn}\in \mathbb{R}^{w \times h \times C}$ be the output feature maps of the two enhancement steps with resolution of $w \times h$ and $C$ channels.
They are first aggregated by element-wise summation into a single feature map, then we divide it into the 4$\times$4 regular grid and conduct average pooling on each cell so that feature vectors for 16 local patches are obtained.
The local feature vectors are then concatenated into one large vector, which is in turn fed to the following fully connected layers.
The final output of the fully connected layers are two score vectors ${\bf s}_\textrm{in} \in \mathbb{R}^{16 C}$ and ${\bf s}_\textrm{bn} \in \mathbb{R}^{16 C}$, each of whose elements indicates how useful a specific channel of $F_\textrm{in}$ or $F_\textrm{bn}$ is for handling the distortion in a specific local patch. 

The score vectors ${\bf s}_\textrm{in}$ and ${\bf s}_\textrm{bn}$ are used to compute attention weights.
They are first reshaped into score tensors $S_\textrm{in} \in \mathbb{R}^{4 \times 4 \times C}$ and $S_\textrm{bn}\in \mathbb{R}^{4 \times 4 \times C}$, then converted to attention weight tensors $A_\textrm{in}$ and $A_\textrm{bn}$ of the same dimension through the element-wise softmax operation:
\begin{align}
    A_\textrm{in}[i,j,c] & = \cfrac{\exp{S_\textrm{in}[i,j,c]}}{\exp{S_\textrm{in}[i,j,c]} + \exp{S_\textrm{bn}[i,j,c]}}, \\
    A_\textrm{bn}[i,j,c] & = \cfrac{\exp{S_\textrm{bn}[i,j,c]}}{\exp{S_\textrm{in}[i,j,c]} + \exp{S_\textrm{bn}[i,j,c]}}.
\end{align}
The spatial dimension of the two attention tensors are enlarged to $w \times h$ by bilinear interpolation to be multiplied to the outputs of the two enhancement steps, $F_\textrm{in}$ and $F_\textrm{bn}$.
The final output of SEM, denoted by $Y \in \mathbb{R}^{w \times h \times C}$, is then computed by weighted summation of $F_\textrm{in}$ and $F_\textrm{bn}$:
\begin{align}
    Y = \big( A'_\textrm{in} \otimes F_\textrm{in} \big) \oplus \big( A'_\textrm{bn} \otimes F_\textrm{bn} \big), \label{eq:integ_w_att}
\end{align}
where $A'_\textrm{in}$ and $A'_\textrm{bn}$ are the resized attention tensors, $\otimes$ indicates element-wise multiplication, and $\oplus$ denotes element-wise summation.

Note that the attention tensors have both spatial (\ie, patch) and channel dimensions. 
In Eq.~\eqref{eq:integ_w_att}, the channel-wise attention facilitates more flexible and diverse integration of the two enhancement steps since different channels can be processed by different enhancement steps; although a SEM provides only two enhancement steps, the number of their output combinations is in principle extremely large.
On the other hand, the patch-wise attention allows to process different image regions with different enhancement processes.



\subsection{Overall Architecture}


\begin{wraptable}{r}{5cm}
\caption{Details of URIE. $W \times H$ denotes the input resolution.}
\centering
\scalebox{0.72}{
\begin{tabular}{@{}C{0.65cm}|@{}C{2.0cm}|@{}C{4.2cm}@{}}
\hline
\multicolumn{2}{c|}{Output} & Layer specification
\\ \hline
(1) & $W \times H$      & $9 \times 9$ conv, 32, stride 1
\\ \hline
(2) & $W/2 \times H/2$  & $2 \times 2$ max pool, stride 2
\\ \hline
(3) & $W/2 \times H/2$  & SEM[$3 \times 3$, 64, stride 1]
\\ \hline
(4) & $W/4 \times H/4$  & $2 \times 2$ max pool, stride 2
\\ \hline
(5) & $W/4 \times H/4$  & SEM[$3 \times 3$, 64, stride 1]
\\ \hline
(6) & $W/2 \times H/2$  & $2 \times$ bilinear upsampling
\\ \hline
(7) & $W/2 \times H/2$  & concat[(3), (6)]
\\ \hline
(8) & $W/2 \times H/2$  & SEM[$3 \times 3$, 32, stride 1]
\\ \hline
(9) & $W \times H$      & $2 \times$ bilinear upsampling
\\ \hline
(10) & $W \times H$     & concat[(5), (9)]
\\ \hline
(11) & $W \times H$     & SEM[$3 \times 3$, 16, stride 1]
\\ \hline
(12) & $W \times H$     & $3 \times 3$ conv, 3, stride 1
\\ \hline
\end{tabular}
}
\label{tab:archi_detail}
\end{wraptable}

The overall architecture of URIE is illustrated in Fig.~\ref{fig:overall_arch}.
Its structure resembles U-Net~\cite{unet}, but is clearly distinct from the conventional U-Net architectures in that its all layers, except for the first and the last, are not ordinary convolution layers but SEMs introduced in Sec.~\ref{sec:selective_enhc}.
The reason for following the design principle of U-Net is two-fold.
First, it effectively enlarges the receptive field so as to capture high-level semantic information.
This is essential for URIE that performs recognition-aware image enhancement, unlike ordinary restoration models that do not necessarily need to understand image contents.
Second, it is computationally efficient since it reduces the sizes of intermediate features using pooling operations.
Reducing feature resolution may degrade the perceptual quality of output as a side effect, which however is not our concern since our objective is improved recognition performance of following recognition model.

The key components in this architecture are SEMs, which allow URIE to handle multiple and latent factors of image degradation efficiently and effectively.
Since each SEM provides two enhancement steps and a sequence of such steps can be regarded as an individual enhancement procedure, the concatenation of SEMs leads to an exponentially large number of enhancement procedures, which can handle a large variety of distortions.
Moreover, through the attentions drawn by SEMs, URIE can dynamically select enhancement procedures depending on the latent type of input distortion.
Since the selection is done for each local patch of the input image through the patch-wise attentions, URIE can handle different image regions with different enhancement procedures.
This is more useful and realistic than applying a single enhancement function to the whole input image since individual regions of the input could be distorted by different factors in realistic scenarios, \eg, images with locally different haze densities with respect to depths, and images partially over- and under-exposed at the same time.

The specification of URIE, including the size of convolution kernels and the number of channels per layer, is given in Tab.~\ref{tab:archi_detail}.
In the table, SEM[$k\times k$, $C$, stride $s$] means that each convolution layer in the SEM consists of $C$ number of $k \times k$ kernels with stride $s$.
Note that we set the size of convolution kernels for the first layer especially large to capture rich information about image distortion and to enlarge the receptive field of our shallow network architecture.

\subsection{Training Strategy}
\label{sec:training_urie}

Our training scheme for URIE is based on two key ideas.
(1) \emph{Recognition-aware loss}:
We employ a pretrained recognition network to provide URIE with supervisory signals that lead it to be recognition-friendly.
Specifically, URIE is coupled with a recognition network that is pretrained on clean images and frozen during training.
It is then trained to minimize a recognition loss of the pretrained model on distorted images in an end-to-end manner.
(2) \emph{Large-scale learning with distorted images}:
We suggest learning URIE using a large-scale image dataset in the presence of diverse distortions so that it may learn enhancement functions insensitive to distortion types and recognition models. 
This is motivated by the fact that deep neural networks trained in large-scale datasets have been used for various purposes other than their original targets (\eg, ImageNet pretrained classifiers used for other recognition tasks~\cite{deconvnet,Oquab2014,vggnet}).

In detail, we employ ResNet-50~\cite{resnet} pretrained on the original ImageNet~\cite{Imagenet} as the coupled recognition model, and adopt the ImageNet-C dataset~\cite{Hendrycks2019_ImageNetC} as our training data. 
URIE is then trained  on the ImageNet-C to minimize the cross-entropy loss of the pretrained model.
The training dataset consists of images from the original ImageNet yet corrupted artificially by in total 19 different distortions with 5 intensity levels.
We use only 15 distortion classes\footnote{Gaussian noise, shot noise, impulse noise, defocus blur, glass blur, motion blur, zoom blur, snow, frost, fog, brightness, contrast, elastic transform, pixelation, jpeg.} among them to degrade training images, and call them \emph{seen corruptions}.
The other distortion classes\footnote{Speckle noise, Gaussian blur, spatter, saturation.} are kept as \emph{unseen corruptions} for evaluating the generalization ability of our model in testing.
Note that clean images from the original ImageNet are also used for training to prevent URIE from being biased to distortions. 

\subsection{Discussion} 
\label{sec:method_discussion}
\begin{figure*} [!t]
\centering
\includegraphics[width = 0.87 \textwidth]{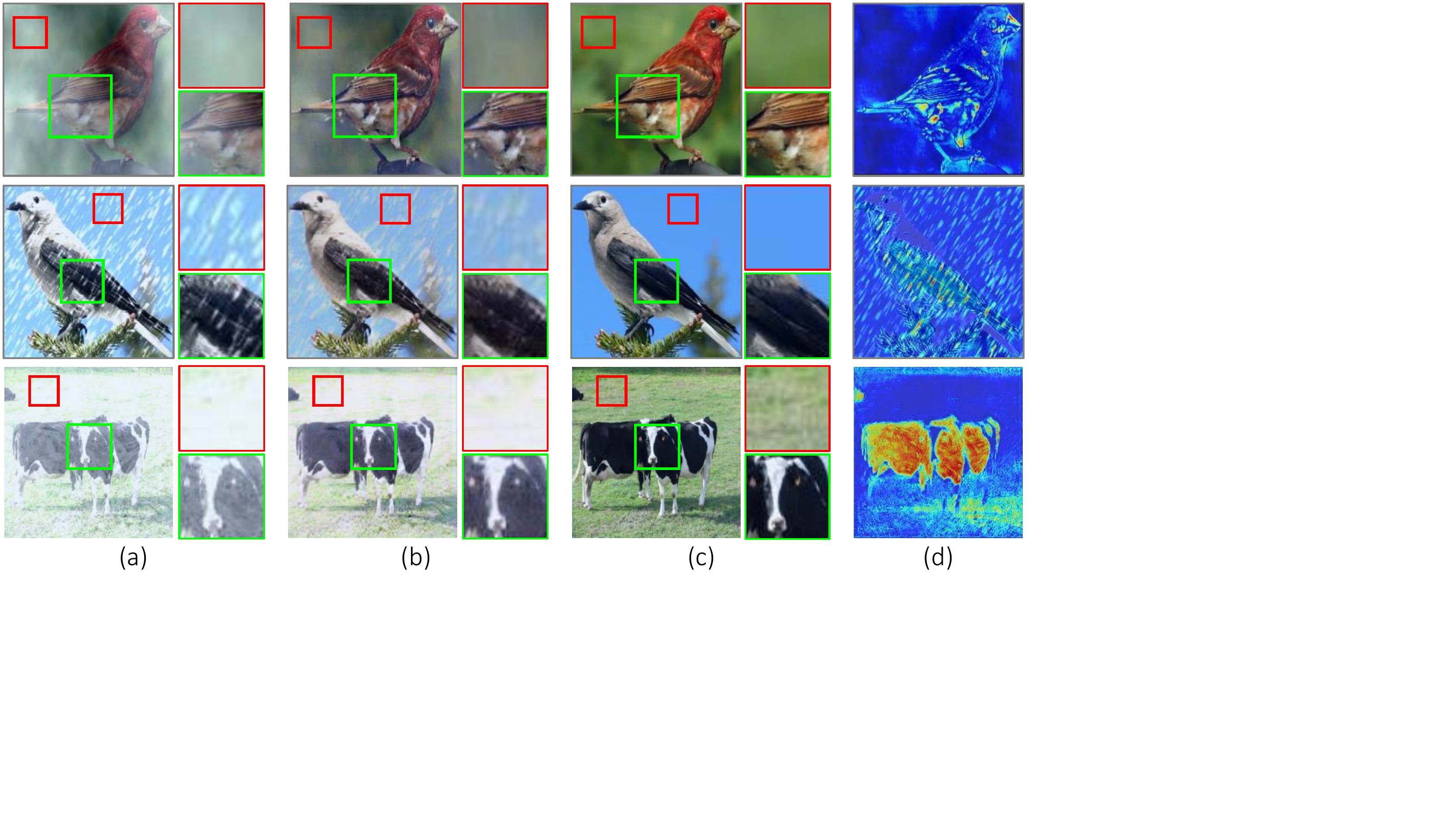}
\caption{Example outputs of URIE.
(a) Distorted input images. 
(b) Outputs of URIE.
(c) Ground-truth images.
(d) Magnitudes of per-pixel intensity change by URIE.
} 
\label{fig:cub_example}
\end{figure*}

\noindent\textbf{How our model works.}
To investigate how URIE works, we apply the model trained in Sec.~\ref{sec:training_urie} to random images sampled from other datasets and corrupted by the same synthetic distortions with the ImageNet-C.
Note that URIE is directly used as-is with no finetuning.
Qualitative results of URIE on these images are given in Fig.~\ref{fig:cub_example}. 
Interestingly, URIE does alleviate the effect of distortions to some degree although it is not trained explicitly for that purpose. 
More importantly, Fig.~\ref{fig:cub_example}(b) and~\ref{fig:cub_example}(d) demonstrate that it tends to manipulate salient local areas only (\eg, object parts) and ignore remainders (\eg, background).
This behavior is a result of the recognition-aware learning in Sec.~\ref{sec:training_urie} that prompts the model to focus more on image regions important for recognition.

\noindent\textbf{Comparison to SKNet.}
URIE shares a similar idea with the Selective Kernel Network (SKNet)~\cite{Li2019_SKNet} that adopts a self-attention mechanism to utilize intermediate feature maps selectively. 
However, the two models are clearly distinct since their overall architectures and training schemes are totally different due to their different objectives: image classification in SKNet, and image enhancement in URIE. 
Besides the above, their attention mechanisms are different in the following aspects.
First, SKNet selects from convolution layers of different kernel sizes to implement dynamic receptive fields, but URIE selects from those with different normalization operations, IN and BN, to adapt dynamically to the latent input distortion.
Second, SKNet draws only channel-wise attentions, but URIE takes both channel and spatial locations into account when drawing attentions to treat different image regions differently.

\noindent\textbf{Comparison to OWAN.}
OWAN~\cite{OWAN_CVPR19} is the current state of the art in image restoration.
It is similar with URIE in that both models consider multiple and latent types of image distortion and use attentions to select appropriate image manipulation process dynamically. 
The major differences between URIE and OWAN lie in their techniques impelementing the ideas. 
OWAN draws attentions on a number of convolution layers to select useful ones among them, thus has to compute individual feature maps of the layers, which is time consuming.
On the other hands, URIE selects from only two enhancement steps, but mix their outputs up in a patch- and channel-wise manner by drawing attention weights on individual channels and local patches.
This scheme allows our model to enjoy an extremely large number of output combinations while keeping computational efficiency.
Thanks to these differences, URIE is substantially more efficient than OWAN, \emph{about 6 times faster} in terms of multiply-accumulate operations.
\section{Experiments}
\label{sec:experiment}


In this section, we empirically verify that URIE improves performance of diverse recognition models for image classification, object detection, and semantic segmentation on real and artificially degraded images.
For image restoration performance of URIE, refer to Tab. 15 of the appendix. 

\subsection{Training Configurations}
Each training image is first resized to 256$\times$256, cropped to 224$\times$224, and flipped horizontally at random for data augmentation.
It is then degraded on-the-fly by the synthetic distortion generator of the ImageNet-C~\cite{Hendrycks2019_ImageNetC}, where the type and intensity of distortion are chosen at random.
Note that only the 15 seen corruptions are used during training as described in Sec.~\ref{sec:training_urie}.
For optimization, we employ Adam~\cite{Adamsolver} with learning rate 0.001, and decay the learning rate every 8 epochs by 10.
Our model is trained for 30 epochs with mini-batches of size 112.

\subsection{Experimental Configurations}
For evaluation of our model, we employ widely used image recognition datasets, ImageNet~\cite{Imagenet}, CUB~\cite{CUB200}, and PASCAL VOC~\cite{Pascalvoc},
and generate three different versions of their test or validation sets:
\emph{Clean}--original uncorrupted images,
\emph{Seen}--images degraded by the 15 seen corruptions, and
\emph{Unseen}--images degraded by the 4 unseen corruptions.
We did not generate the datasets on-the-fly but prior to evaluation so that they are fixed during testing.
In addition, our model is evaluated on the collection of real haze images introduced in~\cite{Pei_2018_ECCV}.

To verify the effectiveness of URIE, it is compared with two different image processing models. 
One of them is OWAN~\cite{OWAN_CVPR19}, the state-of-the-art image restoration model devised to deal with multiple and latent types of image distortion.
OWAN is trained on the same data with ours for fair comparisons, but by using the original loss proposed in~\cite{OWAN_CVPR19} since it is not straightforward to train this model with our recognition-aware loss on the large-scale dataset due to its heavy computational complexity (refer to Sec.~\ref{sec:method_discussion}).
The other one is URIE-MSE, which is a restoration-oriented variant of URIE and trained using a Mean-Squared-Error loss instead of the recognition-aware loss.
URIE and these two models are evaluated in terms of performance of coupled recognizers on the aforementioned datasets.
Note that they are never finetuned to test datasets, and all the recognizers coupled with them are trained on the clean images and tested on the distorted images as-is. 
\begin{table}[!t] \small
\centering

\caption{
Classification accuracy on the ImageNet dataset. The numbers in parentheses indicate the differences from the baseline. V16, R50, and R101 denote VGG-16, ResNet-50, and ResNet-101, respectively.
}
\scalebox{0.7}{
\begin{tabular}{@{}C{1cm}|@{}C{1.8cm}@{}C{1.8cm}@{}C{1.8cm}|@{}C{1.8cm}@{}C{1.8cm}@{}C{1.8cm}|@{}C{1.8cm}@{}C{1.8cm}@{}C{1.8cm}@{}}
\multirow{2}{*}{}   & \multicolumn{3}{c|}{OWAN}  & \multicolumn{3}{c|}{URIE-MSE} & \multicolumn{3}{c}{URIE}  \\ \cline{2-10} & \multicolumn{1}{c}{Clean} & \multicolumn{1}{c}{Seen} & \multicolumn{1}{c|}{Unseen} & \multicolumn{1}{c}{Clean} & \multicolumn{1}{c}{Seen} & \multicolumn{1}{c|}{Unseen} & \multicolumn{1}{c}{Clean} & \multicolumn{1}{c}{Seen} & \multicolumn{1}{c}{Unseen} \\ \hline
V16 & 69.8 \red{(-0.2)} & 31.6 \green{(+2.0)} & 39.8 \green{(+0.8)} & 59.5 \red{(-10.5)} & 32.3 \green{(+2.7)} & 38.1 \red{(-0.9)} & 67.1 \red{(-2.9)} & 42.4 \green{(+12.8)} & 44.8 \green{(+5.8)} \\
R50 & 74.4 \red{(-0.4)} &  42.6 \green{(+2.0)} & 50.3 \green{(+1.2)} & 66.7 \red{(- 7.8)} & 44.5 \green{(+3.9)} & 49.3 \green{(+0.2)} & 72.9 \red{(-1.6)} & 55.1 \green{(+14.5)} & 56.5 \green{(+7.4)}  \\
R101& 75.9 ( 0.0) & 47.4 \green{(+1.5)} & 54.6 \green{(+0.8)} &  68.8 \red{(- 7.1)} & 50.0 \green{(+4.1)} & 54.0 \green{(+0.2)} & 74.1 \red{(-1.8)} & 57.8 \green{(+11.9)} & 59.4 \green{(+5.6)} 
\end{tabular}
}
\label{tab:classification_on_imagenet}
\end{table}

\begin{table}[!t] \small
\centering
\caption{Classification accuracy on the CUB dataset. The numbers in parentheses indicate the differences from the baseline. V16, R50, and R101 denote VGG-16, ResNet-50, and ResNet-101, respectively.
}
\scalebox{0.7}{
\begin{tabular}{@{}C{1cm}|@{}C{1.8cm}@{}C{1.8cm}@{}C{1.8cm}|@{}C{1.8cm}@{}C{1.8cm}@{}C{1.8cm}|@{}C{1.8cm}@{}C{1.8cm}@{}C{1.8cm}@{}}
\multirow{2}{*}{}   & \multicolumn{3}{c|}{OWAN}  & \multicolumn{3}{c|}{URIE-MSE} & \multicolumn{3}{c}{URIE}  \\ \cline{2-10} & \multicolumn{1}{c}{Clean} & \multicolumn{1}{c}{Seen} & \multicolumn{1}{c|}{Unseen} & \multicolumn{1}{c}{Clean} & \multicolumn{1}{c}{Seen} & \multicolumn{1}{c|}{Unseen} & \multicolumn{1}{c}{Clean} & \multicolumn{1}{c}{Seen} & \multicolumn{1}{c}{Unseen} \\ \hline
V16 & 79.1 ( 0.0) & 50.4 \green{(+4.2)} & 42.3 \green{(+1.7)} & 68.5 \red{(-11.6)} & 50.5 \green{(+4.3)} & 39.6 \red{(-1.0)} & 77.9 \red{(-1.2)} & 58.8 \green{(+12.6)} & 48.9 \green{(+8.3)} \\
R50 & 84.6 ( 0.0) & 52.8 \green{(+3.3)} & 48.2 \green{(+1.2)} & 76.5 \red{(- 8.1)} & 58.4 \green{(+8.9)} & 49.3 \green{(+2.1)} & 83.7 \red{(-0.9)} & 64.7 \green{(+15.2)} & 54.9 \green{(+7.7)}  \\
R101& 84.9 \red{(-0.1)} & 55.5 \green{(+3.6)} & 48.5 \green{(+1.8)} & 76.2 \red{(- 8.8)} & 59.9 \green{(+8.0)} & 50.5 \green{(+3.8)} & 84.1 \red{(-0.9)} & 67.1 \green{(+15.2)} & 56.6 \green{(+9.9)}
\end{tabular}
}
\label{tab:classificaion_on_cub}
\end{table}

\begin{figure*} [!t]
\centering
\includegraphics[width = 1 \textwidth]{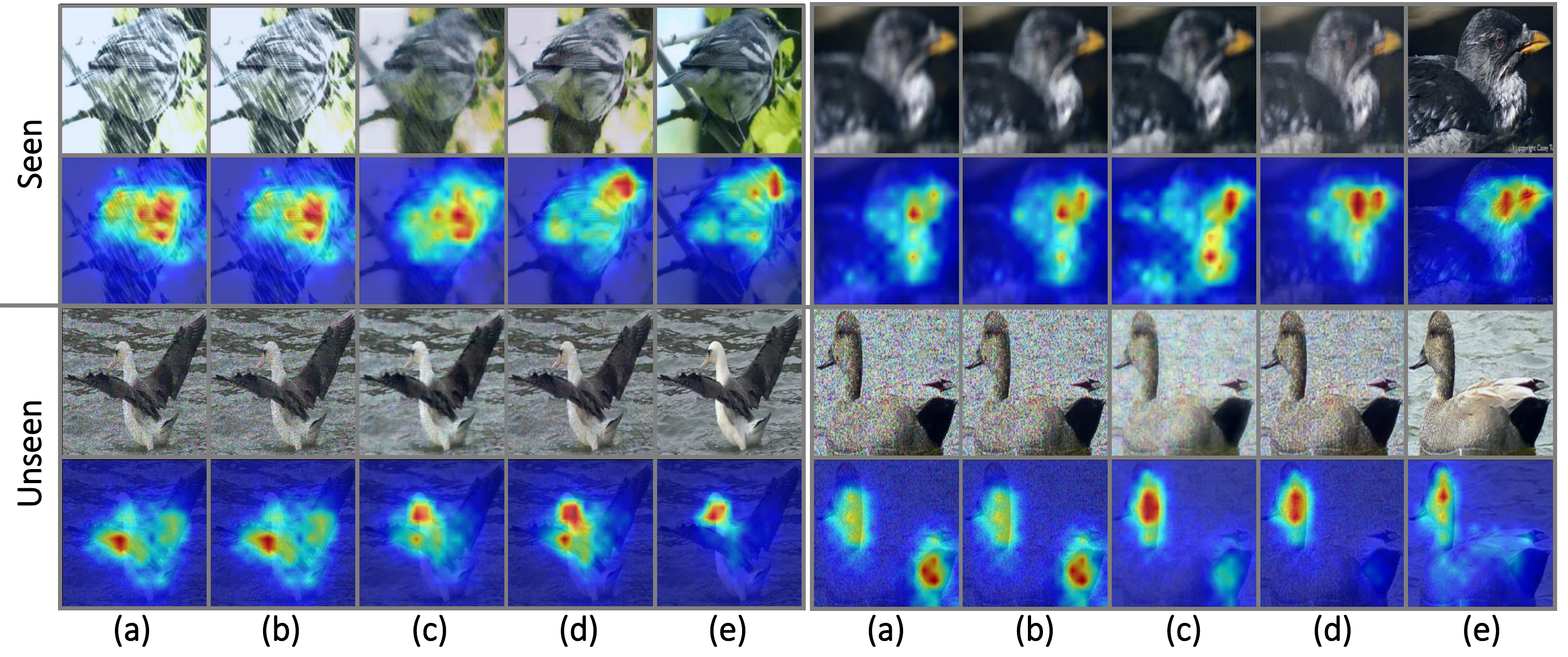}
\caption{
Qualitative results on the CUB dataset. 
(a) Input distorted images.
(b) OWAN.
(c) URIE-MSE.
(d) URIE.
(e) Ground-truth images.
For all images, their grad-CAMs drawn by the ResNet-50 classifier are presented alongside.
Examples in the first row are degraded by seen corruptions and the others are by unseen corruptions.
}
\label{fig:motiv}
\end{figure*}

\subsection{Performance Evaluation}
\noindent\textbf{Classification on ImageNet.}
We first evaluate and compare the effectiveness of URIE and the other two models for classification on the ImageNet validation set.
For evaluation, they are integrated with three different classification networks: VGG-16~\cite{vggnet}, ResNet-50~\cite{resnet} and ResNet-101~\cite{resnet}.
As can be seen in Tab.~\ref{tab:classification_on_imagenet}, the performance improvement by URIE is substantial in the case of seen corruptions and nontrivial for unseen corruptions as well, regardless of the classification network it is integrated with.
These results suggest that URIE is universally applicable to diverse corruptions and to various recognition models.
Also, the performance gap between URIE and the other two methods is significant, which verifies the advantage of our method. 
In particular, the difference between URIE and URIE-MSE in performance shows the clear advantage of our recognition-aware training.
Moreover, URIE-MSE is comparable to OWAN, although it is six times more efficient as discussed in Sec.~\ref{sec:method_discussion}. 
This demonstrates the advantage of the network architecture we proposed. 
Unfortunately, URIE marginally degrades the performance in the case of \emph{clean}; we suspect that the architecture of URIE could be biased to distortions while OWAN has more skip connections and better preserves input consequently.
\begin{table}[!t] \small
\centering
\caption{Object detection performance of SSD~300 in mAP (\%) on the VOC 2007 dataset. The numbers in parentheses indicate the differences from the baseline.
}
\scalebox{0.74}{
\begin{tabular}{@{}C{1.8cm}@{}C{1.8cm}@{}C{1.8cm}|@{}C{1.8cm}@{}C{1.8cm}@{}C{1.8cm}|@{}C{1.8cm}@{}C{1.8cm}@{}C{1.8cm}@{}}
\multicolumn{3}{c|}{OWAN}  & \multicolumn{3}{c|}{URIE-MSE} & \multicolumn{3}{c}{URIE} \\
\hline
\multicolumn{1}{c}{Clean} & \multicolumn{1}{c}{Seen} & \multicolumn{1}{c|}{Unseen} & \multicolumn{1}{c}{Clean} & \multicolumn{1}{c}{Seen} & \multicolumn{1}{c|}{Unseen} & \multicolumn{1}{c}{Clean} & \multicolumn{1}{c}{Seen} & \multicolumn{1}{c}{Unseen} \\ 
\hline
77.4 (0.0) &  51.3 \green{(+1.7)}   & 60.2 \green{(+0.9)} & 76.5 \red{(-0.9)} & 52.7 \green{(+3.1)} & 59.9 \green{(+0.6)} & 76.5 \red{(-0.9)} & 59.4 \green{(+9.8)} & 62.7 \green{(+3.4)} \\
\end{tabular}
}
\label{tab:detection_on_voc}
\end{table}

\begin{figure*} [!t]
\centering
\includegraphics[width = 1 \textwidth]{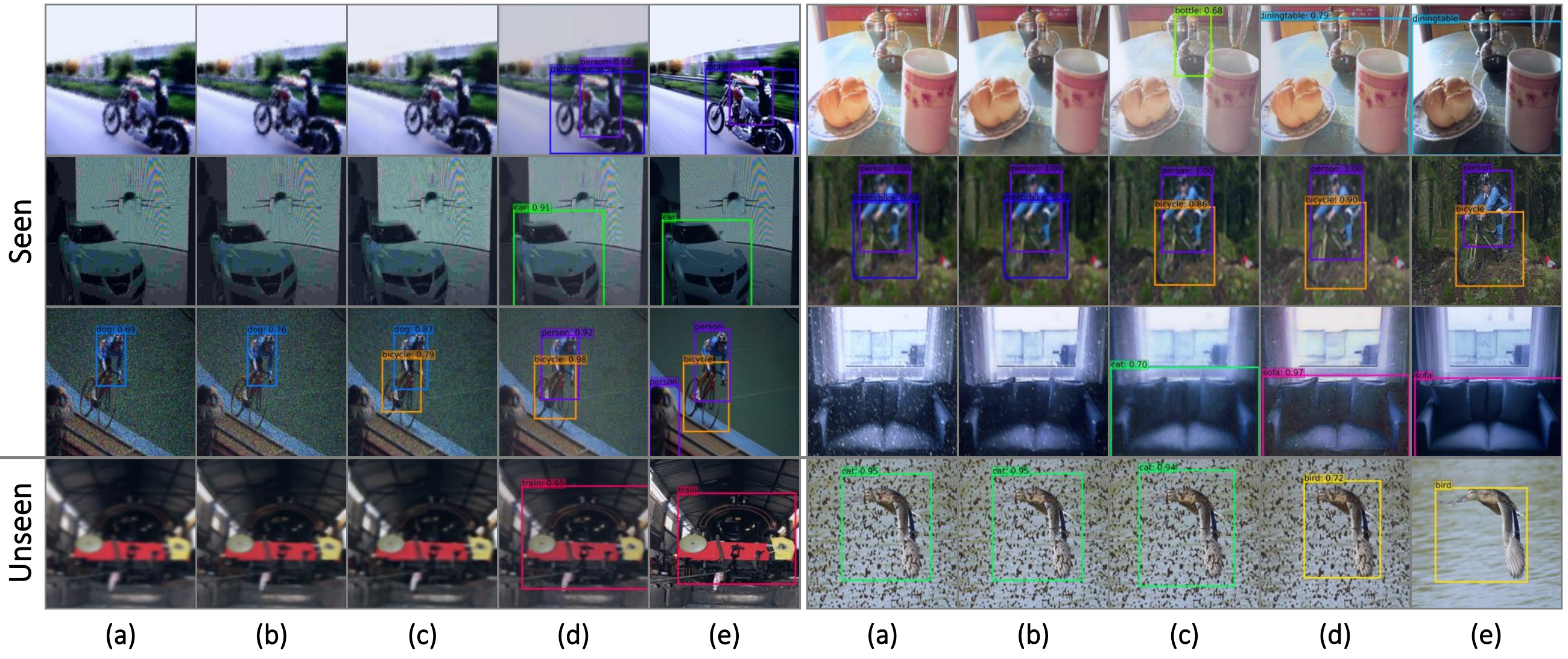}
\caption{
Qualitative results of SSD~300 on the VOC 2007 dataset. 
(a) Corrupted input. 
(b) OWAN. 
(c) URIE-MSE. 
(d) URIE. 
(e) Ground-truth.
Examples in the top three rows are degraded by seen corruptions and the others are by unseen corruptions.
} 
\label{fig:voc_det}
\end{figure*}

\noindent\textbf{Fine-grained classification on CUB.}
Our model and the others are evaluated on the CUB validation set to examine 
if they work universally even in other domains where data distributions are different from that of the training data, \ie, ImageNet.
The evaluation setting in this dataset is the same with that of the ImageNet classification.
As summarized in Tab.~\ref{tab:classificaion_on_cub}, all models turn out to be transferable to some degree in spite of the domain gap between the ImageNet and CUB datasets; we suspect that this is an effect of the large-scale training using the ImageNet-C.
However, URIE demonstrates its superiority over the other models by outperforming them substantially.
Fig.~\ref{fig:cub_example} presents qualitative results of the models and their grad-CAMs~\cite{grad-cam} computed by the ResNet-50 classifier using the ground-truth labels.
In this figure, grad-CAMs computed from the results of URIE are closest to those of ground-truth images, which means that URIE best recovers image regions critical for correct classification.

\begin{figure*} [!t]
\centering
\includegraphics[width = 1 \textwidth]{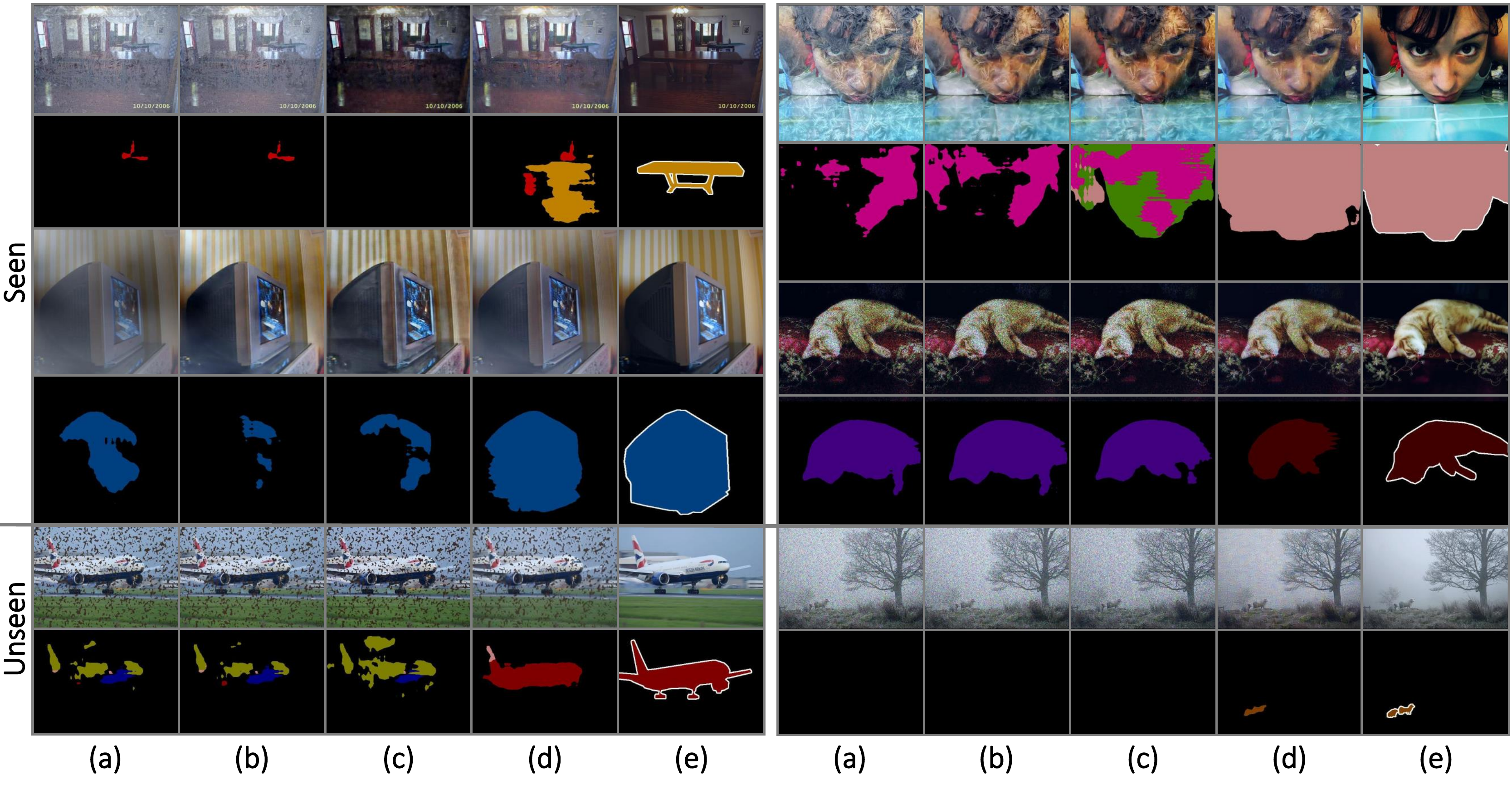}
\caption{
Qualitative results of DeepLab~v3 on the VOC 2012 dataset. 
(a) Corrupted input. 
(b) OWAN. 
(c) URIE-MSE. 
(d) URIE. 
(e) Ground-truth.
Examples in the top two rows are degraded by seen corruptions and the others are by unseen corruptions.
} 
\label{fig:voc_seg}
\end{figure*}

\noindent\textbf{Object detection and segmentation on PASCAL VOC.}
This time we evaluate the effectiveness of our model for object detection and semantic segmentation on the PASCAL VOC dataset to validate its universal applicability to totally different recognition tasks. 
We employ SSD~300~\cite{SSD} for object detection and DeepLab~v3~\cite{deeplab_v3} for semantic segmentation as the recognition models. 
Following the convention, the object detection model is evaluated on the VOC 2007 test set while the segmentation model is tested on the VOC 2012 validation set. 
Tab.~\ref{tab:detection_on_voc} and~\ref{tab:segmentation_on_voc} summarize performance of the two recognition models, and Fig.~\ref{fig:voc_det} and~\ref{fig:voc_seg} present their qualitative results when coupled with URIE and the two restoration models. 
In both of object detection and semantic segmentation, URIE improves recognition performance noticeably, and clearly outperforms the other two models for both seen and unseen corruptions.

\begin{figure*} [!t]
\centering
\includegraphics[width = 1 \textwidth]{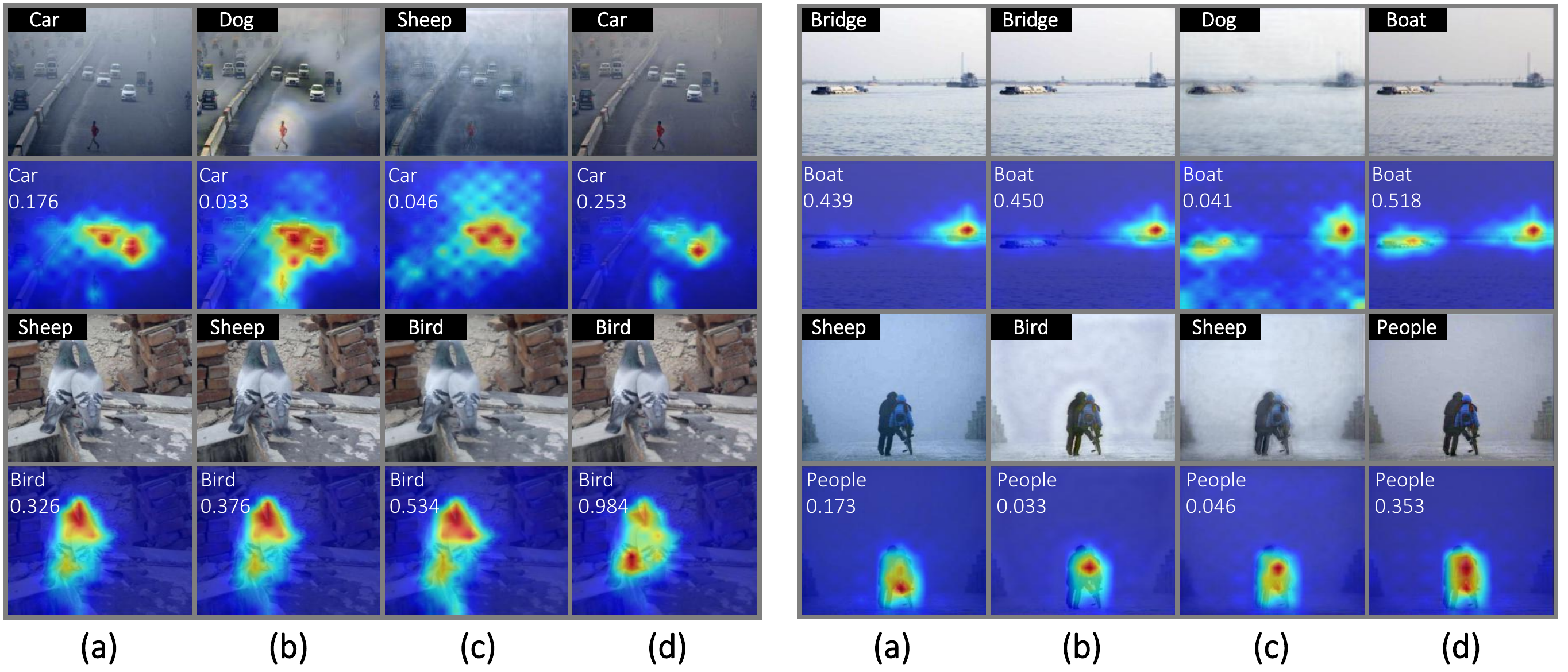}
\caption{
Qualitative results on the Haze-20 dataset.
(a) Corrupted input. 
(b) OWAN. 
(c) URIE-MSE. 
(d) URIE. 
Top-1 prediction of the ResNet-50 classifier together with its confidence score and grad-CAM are presented alongside per example.
} 
\label{fig:haze}
\end{figure*}

\begin{table}[!t] \small
\centering
\caption{
Semantic segmentation performance of DeepLab~v3 in mIoU (\%) on the VOC 2012 dataset. The numbers in parentheses indicate the differences from the baseline.
}
\scalebox{0.74}{
\begin{tabular}{@{}C{1.8cm}@{}C{1.8cm}@{}C{1.8cm}|@{}C{1.8cm}@{}C{1.8cm}@{}C{1.8cm}|@{}C{1.8cm}@{}C{1.8cm}@{}C{1.8cm}@{}}
\multicolumn{3}{c|}{OWAN}  & \multicolumn{3}{c|}{URIE-MSE} & \multicolumn{3}{c}{URIE}  \\
\hline
\multicolumn{1}{c}{Clean} & \multicolumn{1}{c}{Seen} & \multicolumn{1}{c|}{Unseen} & \multicolumn{1}{c}{Clean} & \multicolumn{1}{c}{Seen} & \multicolumn{1}{c|}{Unseen} & \multicolumn{1}{c}{Clean} & \multicolumn{1}{c}{Seen} & \multicolumn{1}{c}{Unseen} \\ 
\hline
78.7 \red{(-0.2)} &  58.2 \green{(+3.3)}   &  65.4\green{(+2.1)} & 78.1 \red{(-0.8)} & 60.1 \green{(+5.2)} & 63.6 \green{(+0.3)} & 78.6 \red{(-0.3)} & 67.0 \green{(+12.1)} & 67.2 \green{(+3.9)} \\
\end{tabular}
}
\label{tab:segmentation_on_voc}
\end{table}
\noindent\textbf{Classification on Haze-20.}
Finally, the effectiveness of URIE is evaluated on the Haze-20 dataset~\cite{Pei_2018_ECCV}, containing haze images of 20 object classes captured in real environments. 
This dataset is paired with another set, HazeClear-20, a collection of real haze-free images of the same 20 object categories. 
URIE and the two restoration models are applied to both of datasets and evaluated in terms of performance of a ResNet-50 classifier taking their results as input.
This setting is challenging for the three models since they are trained only with synthetically distorted images. 
Moreover, Pei~\etal~\cite{Pei_2018_ECCV} demonstrated by extensive experiments that existing dehazing algorithms do not help improve classification performance on the Haze-20 dataset.
Nevertheless, as summarized in Tab.~\ref{tab:classification_on_haze}, URIE is still able to improve the classification accuracy in the presence of real-world haze, and at the same time, outperforms the two restoration methods.
Fig.~\ref{fig:haze} shows that URIE enhances areas around objects better than OWAN and URIE-MSE, and consequently allows the coupled classifier to predict the correct class label and draw attention most accurately.

\begin{table}[!t] \small
\centering
\caption{Accuracy of the ResNet-50 classifier on the Haze-20 and HazeClear-20 datasets.
The numbers in parentheses indicate the differences from the baseline.
}
\scalebox{0.9}{
\begin{tabular}{@{}C{1.8cm}@{}C{1.8cm}|@{}C{1.8cm}@{}C{1.8cm}|@{}C{1.8cm}@{}C{1.8cm}}
\multicolumn{2}{c|}{OWAN}  & \multicolumn{2}{c|}{URIE-MSE} & \multicolumn{2}{c}{URIE}  \\ \hline
\multicolumn{1}{c}{Haze} & \multicolumn{1}{c|}{HazeClear} & \multicolumn{1}{c}{Haze} & \multicolumn{1}{c|}{HazeClear} & \multicolumn{1}{c}{Haze} & \multicolumn{1}{c}{HazeClear} \\ \hline
 58.7 \green{(+1.7)} &  97.8 \red{(-0.1)} &  53.4 \red{(-3.6)} & 94.4 \red{(-3.5)} & 60.8 \green{(+3.8)} &  97.6 \red{(-0.3)} \\
\end{tabular}
}

\label{tab:classification_on_haze}
\end{table}

\section{Conclusion}
\label{sec:conclusion}

This paper has presented a new image enhancement network dedicated to robust visual recognition in the presence of input distortion.
Our network is universally applicable to various types of image distortions and many different recognition models.
At the same time, it is recognition-friendly since it is optimized to improve robustness of recognition models taking its output as input.
These features of our model are given by a novel network architecture and a training strategy we carefully designed.
The advantages of our model have been verified throughout experiments in various settings.
We however have found that our model marginally degrades performance on distortion-free images. 
Next on our agenda is to resolve this issue by revising the model architecture.

\bigskip
{\small
\noindent \textbf{Acknowledgement:} This work was supported by Samsung Research Funding \& Incubation Center
of Samsung Electronics under Project Number SRFC-IT1801-05.
}

\clearpage
%
%
\bibliographystyle{splncs04}
\bibliography{cvlab_kwak}

\clearpage

\begin{center}
\textbf{\large Appendix}
\end{center}

\setcounter{section}{0}

\renewcommand*{\thesection}{\Alph{section}.}

This appendix presents experimental results omitted from the main paper due to the space limit. 
\Sec{abliation_norm} empirically justifies the combination of Batch Normalization (BN)~\cite{Batchnorm} and Instance Normalization (IN)~\cite{Instancenorm} in our Selective Enhancement Module (SEM). 
In~\Sec{largescale}, we verify the impact of the large-scale learning in terms of performance and universality of URIE by examining the same models trained with smaller datasets.
Also, \Sec{comparison_with_more_baselines} explains why it is not straightforward to compare URIE with existing image restoration models in our setting for robust visual recognition,
and \Sec{restoration_performance} analyzes image restoration performance of URIE. 
Finally, \Sec{qual} presents more qualitative results of object detection and semantic segmentation on the distorted PASCAL VOC datasets.

\section{Ablation Study Regarding Different Normalizations}
\label{sec:abliation_norm}

The two enhancement steps in SEM have different normalization operations, IN and BN.
This section demonstrates that the combination of IN and BN comes with benefits due to their complementary roles in image enhancement.
To this end, we design two variants of URIE, \emph{URIE-BN} and \emph{URIE-IN}, and compare them with the original one.
Specifically, in URIE-BN all normalization operations are BN, and in URIE-IN those are all implemented as IN.
For a fair comparison, the two models are trained in the same setting with URIE.

\begin{figure*} [!t]
\centering
\includegraphics[width = 1 \textwidth] {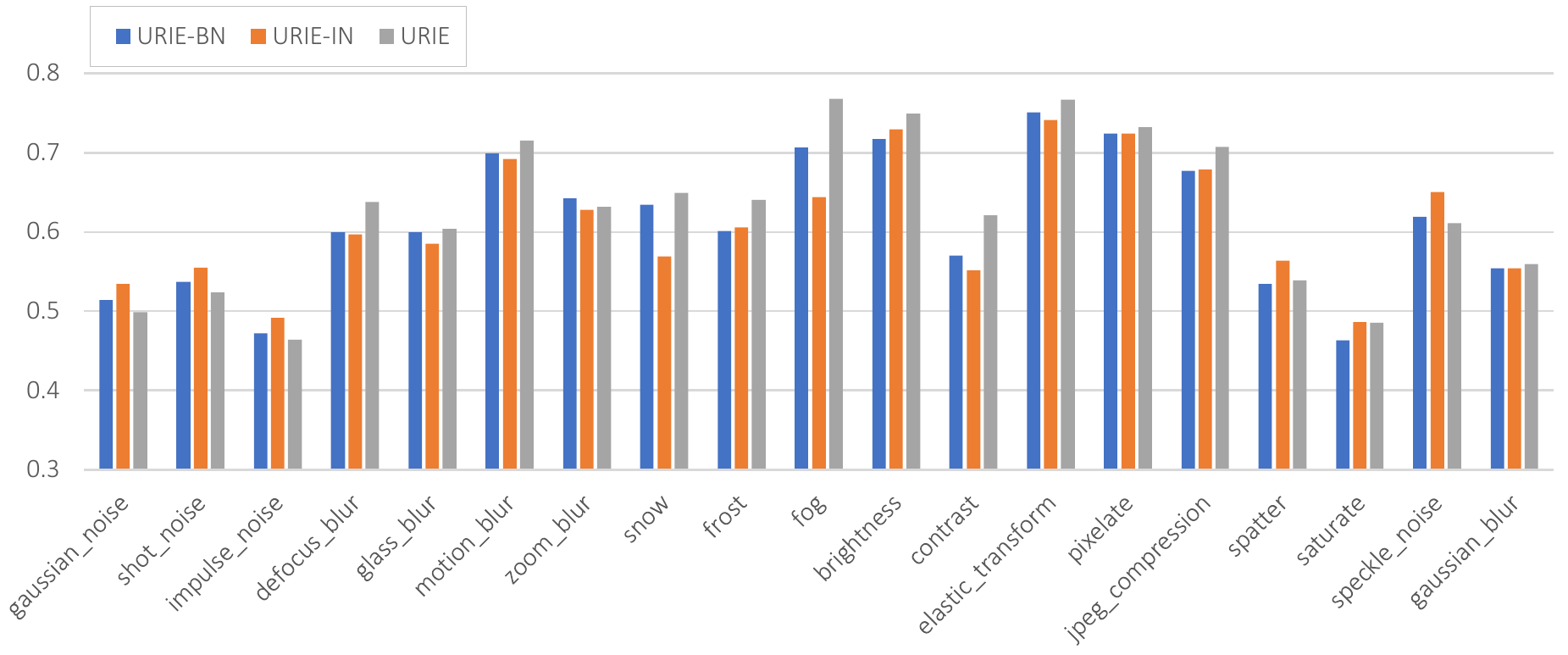}
\caption{
Performance comparison between  URIE-BN, URIE-IN, and URIE per distortion type on the corrupted CUB dataset.
}
\label{fig:corruption_wise_accuracy}
\end{figure*}

The three models are evaluated in terms of recognition performance on distorted images by following the protocol in the main paper.
Tab.~\ref{tab:classificaion_on_imagenet_supp},~\ref{tab:classificaion_on_cub_supp},~\ref{tab:detection_on_voc_supp},~and~\ref{tab:segmentation_on_voc_supp} summarize their performance on the four different recognition tasks.
These results show that using only one of the normalization operations degrades the final recognition performance noticeably in almost all settings.
To investigate the effect of IN and BN in more detail, we measure the performance of the three models per distortion type on the distorted CUB dataset. 
As shown in Fig.~\ref{fig:corruption_wise_accuracy}, URIE-BN and URIE-IN exhibit clearly different tendencies.
URIE-IN works better than URIE-BN for noise-type distortions while URIE-BN dominates URIE-IN when images are corrupted by blur-type distortions or adverse weathers. 
On the other hand, URIE adopting both of IN and BN performs best for most of the distortion types. 
These results justify our assumption that the two different normalization operations are complementary to each other.

\section{Impact of Large Scale Training}
\label{sec:largescale}

To demonstrate the advantage of large-scale training using the ImageNet dataset, we compare URIE with its other two variants,
\emph{URIE-1/4} and \emph{URIE-1/16}, trained using a quarter and a sixteenth of the corrupted ImageNet dataset, respectively. 
Tab.~\ref{tab:classificaion_on_imagenet_quarter},~\ref{tab:classificaion_on_cub_quarter},~\ref{tab:detection_on_voc_quarter},~and~\ref{tab:segmentation_on_voc_quarter} summarize the performance of the three models, and show that URIE-1/4 and URIE-1/16 degrade the recognition performance substantially and are not well transferred to other tasks compared to the original URIE. 
These results justify our assumption that training on a large-scale distorted image dataset can improve the performance and universality of URIE. 


\begin{table}[!t] \small
\centering
\caption{
Classification accuracy on the ImageNet dataset. 
The numbers in parentheses indicate the differences from the performance of URIE. 
V16, R50, and R101 denote VGG-16, ResNet-50, and ResNet-101, respectively.
}
\scalebox{0.75}{
\begin{tabular}{@{}C{1cm}|@{}C{1.3cm}@{}C{1.3cm}@{}C{1.3cm}|@{}C{1.8cm}@{}C{1.8cm}@{}C{1.8cm}@{}|@{}C{1.8cm}@{}C{1.8cm}@{}C{1.8cm}@{}}
\multirow{2}{*}{}   & \multicolumn{3}{c|}{URIE}  & \multicolumn{3}{c|}{URIE-BN} & \multicolumn{3}{c}{URIE-IN} \\ 
\cline{2-10} & \multicolumn{1}{c}{Clean} & \multicolumn{1}{c}{Seen} & \multicolumn{1}{c|}{Unseen} & \multicolumn{1}{c}{Clean} & \multicolumn{1}{c}{Seen} & \multicolumn{1}{c|}{Unseen} & \multicolumn{1}{c}{Clean} & \multicolumn{1}{c}{Seen} & \multicolumn{1}{c}{Unseen}\\ \hline
V16 & 67.1 & 42.4 & 44.8 & 63.0 \red{(-4.1)} & 41.3 \red{(-1.1)} & 43.5 \red{(-1.3)} & 63.9 \red{(-3.2)} & 39.9 \red{(-2.5)} & 44.4 \red{(-0.4)}\\
R50 & 72.9 & 55.1 & 56.5 & 70.4 \red{(-2.5)} & 54.0 \red{(-1.1)} & 55.0 \red{(-1.5)} & 71.0 \red{(-1.9)} & 52.8 \red{(-2.3)} & 55.6 \red{(-0.9)}\\
R101& 74.1 & 57.8 & 59.4 & 71.7 \red{(-2.4)} & 56.8 \red{(-1.0)} & 58.1 \red{(-1.3)} & 72.1 \red{(-2.0)} & 55.5 \red{(-2.3)} & 58.1 \red{(-1.3)}
\end{tabular}
}
\label{tab:classificaion_on_imagenet_supp}
\end{table}

%
%
%

\begin{table}[!t] \small
\centering
\caption{
Classification accuracy on the CUB dataset. 
The numbers in parentheses indicate the differences from the performance of URIE. 
V16, R50, and R101 denote VGG-16, ResNet-50, and ResNet-101, respectively.
}
\scalebox{0.75}{
\begin{tabular}{@{}C{1cm}|@{}C{1.3cm}@{}C{1.3cm}@{}C{1.3cm}|@{}C{1.8cm}@{}C{1.8cm}@{}C{1.8cm}@{}|@{}C{1.8cm}@{}C{1.8cm}@{}C{1.8cm}@{}}
\multirow{2}{*}{}   & \multicolumn{3}{c|}{URIE}  & \multicolumn{3}{c|}{URIE-BN} & \multicolumn{3}{c}{URIE-IN} \\ 
\cline{2-10} & \multicolumn{1}{c}{Clean} & \multicolumn{1}{c}{Seen} & \multicolumn{1}{c|}{Unseen} & \multicolumn{1}{c}{Clean} & \multicolumn{1}{c}{Seen} & \multicolumn{1}{c|}{Unseen} & \multicolumn{1}{c}{Clean} & \multicolumn{1}{c}{Seen} & \multicolumn{1}{c}{Unseen} \\ \hline
V16 & 77.9 & 58.8 & 48.9 & 76.7 \red{(-1.3)} & 56.2 \red{(-2.2)} & 47.4 \red{(-1.5)} & 77.0 \red{(-0.9)} & 56.8 \red{(-2.0)} & 49.8 \green{(+0.9)}  \\
R50 & 83.7 & 64.7 & 54.9 & 82.4 \red{(-1.3)} & 62.9 \red{(-1.8)} & 54.3 \red{(-0.6)} & 82.9 \red{(-0.8)} & 62.1 \red{(-2.6)} & 56.4 \green{(+1.5)}  \\
R101& 84.1 & 67.1 & 56.6 & 82.9 \red{(-1.2)} & 65.0 \red{(-2.1)} & 57.0 \green{(+0.4)} & 82.6 \red{(-1.5)} & 64.8 \red{(-2.3)} & 57.4 \green{(+0.8)} 
\end{tabular}
}
\label{tab:classificaion_on_cub_supp}
\end{table}

%
%
%

\begin{table}[!t] \small
\centering
\caption{
Object detection performance of SSD 300 in mAP (\%) on the VOC 2007 dataset. 
The numbers in parentheses indicate the differences from the scores of URIE.
}
\scalebox{0.75}{
\begin{tabular}{@{}C{1.3cm}@{}C{1.3cm}@{}C{1.3cm}|@{}C{1.8cm}@{}C{1.8cm}@{}C{1.8cm}@{}|@{}C{1.8cm}@{}C{1.8cm}@{}C{1.8cm}@{}}
\multicolumn{3}{c|}{URIE}  & \multicolumn{3}{c|}{URIE-BN} & \multicolumn{3}{c}{URIE-IN}  \\ \hline
\multicolumn{1}{c}{Clean} & \multicolumn{1}{c}{Seen} & \multicolumn{1}{c|}{Unseen} & \multicolumn{1}{c}{Clean} & \multicolumn{1}{c}{Seen} & \multicolumn{1}{c|}{Unseen} & \multicolumn{1}{c}{Clean} & \multicolumn{1}{c}{Seen} & \multicolumn{1}{c}{Unseen} \\ \hline
 76.5 & 59.4 & 62.7 & 74.4 \red{(-2.1)} & 57.9 \red{(-1.5)} & 61.0 \red{(-1.7)} & 74.1 \red{(-2.4)} & 56.9 \red{(-2.5)} & 60.9 \red{(-1.8)} \\
\end{tabular}
}
\label{tab:detection_on_voc_supp}
\end{table}

\begin{table}[!t] \small
\centering
\caption{
Semantic segmentation performance of DeepLab v3 in mIoU (\%) on the VOC 2012 dataset.
The numbers in parentheses indicate the differences from the score of URIE.
}
\scalebox{0.75}{
\begin{tabular}{@{}C{1.3cm}@{}C{1.3cm}@{}C{1.3cm}|@{}C{1.8cm}@{}C{1.8cm}@{}C{1.8cm}@{}|@{}C{1.8cm}@{}C{1.8cm}@{}C{1.8cm}@{}}
\multicolumn{3}{c|}{URIE}  & \multicolumn{3}{c|}{URIE-BN} & \multicolumn{3}{c}{URIE-IN}  \\ \hline
\multicolumn{1}{c}{Clean} & \multicolumn{1}{c}{Seen} & \multicolumn{1}{c|}{Unseen} & \multicolumn{1}{c}{Clean} & \multicolumn{1}{c}{Seen} & \multicolumn{1}{c|}{Unseen} & \multicolumn{1}{c}{Clean} & \multicolumn{1}{c}{Seen} & \multicolumn{1}{c}{Unseen} \\ \hline
 78.6 & 67.0 & 67.2 & 78.4 \red{(-0.2)} & 66.3 \red{(-0.7)} & 68.0 \green{(+0.8)} & 78.5 \red{(-0.1)} & 64.9 \red{(-2.1)} & 67.4 \green{(+0.2)}  \\
\end{tabular}
}
\label{tab:segmentation_on_voc_supp}
\end{table}

\section{Practical Issues on Direct Comparisons to Image Restoration Models}
\label{sec:comparison_with_more_baselines}
To prove effectiveness of proposed method, it would be best to compare URIE to image restoration models trained in the same manner with URIE. 
However, we would stress that it is often impractical to train and evaluate them in the same setting. 
In our setting, specifically, URIE is trained with the recognition-aware loss on the ImageNet-C dataset whose images are corrupted by diverse and latent distortions. 
Hence models must be (1) efficient in computation and memory usage and (2) able to deal with a multitude of latent distortion types. 
Unfortunately, most prior studies on image restoration do not meet the two conditions since they rely on considerably heavier networks and assume a single distortion type. 
In particular, we found that it takes impractically long time to train such enhancement networks with the recognition-aware loss on the ImageNet-C dataset (\eg, taking 137 days on 4 Tesla P40 GPUs in the case of OWAN~\cite{OWAN_CVPR19}).

\section{Restoration Performance of URIE and Its Variants}
\label{sec:restoration_performance}
This section presents restoration performance of URIE and its variants.
They are evaluated in terms Mean Squared Error (MSE) and Structural SIMilarity (SSIM). 
In addition to three models used in our experiment, we consider another variant of URIE that is trained with SSIM loss, called URIE-SSIM.

As reported in the Tab.~\ref{tab:restoration_performance}, URIE is worse than URIE-MSE and URIE-SSIM in restoration but substantially outperforms them in recognition, which suggests that URIE works as desired. 
Also, URIE and its variants outperform OWAN~\cite{OWAN_CVPR19} in recognition performance. 
This is partly due to the superiority of our network architecture, which better handles diverse distortions and images captured in uncontrolled environments.

\section{More Qualitative Results}
\label{sec:qual}

\begin{figure*} [!t]
\centering
\includegraphics[width = 1 \textwidth] {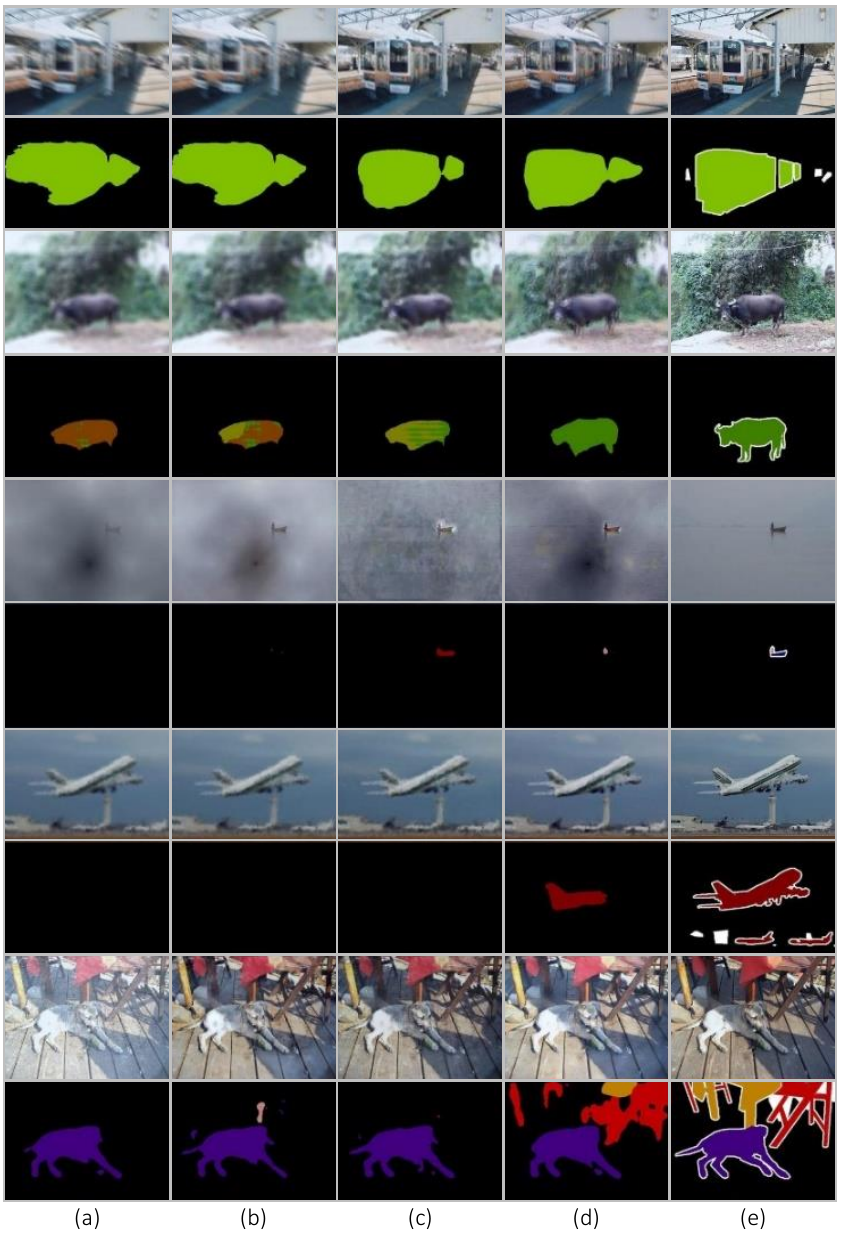}
\caption{
Additional qualitative results of DeepLab v3 on the VOC 2012 dataset. 
(a) Corrupted input. 
(b) OWAN~\cite{OWAN_CVPR19}. 
(c) URIE-MSE. 
(d) URIE. 
(e) Ground-truth. 
}
\vspace{-1mm}
\label{fig:seg_1}
\end{figure*}

\begin{figure*} [!t]
\centering
\includegraphics[width = 1 \textwidth] {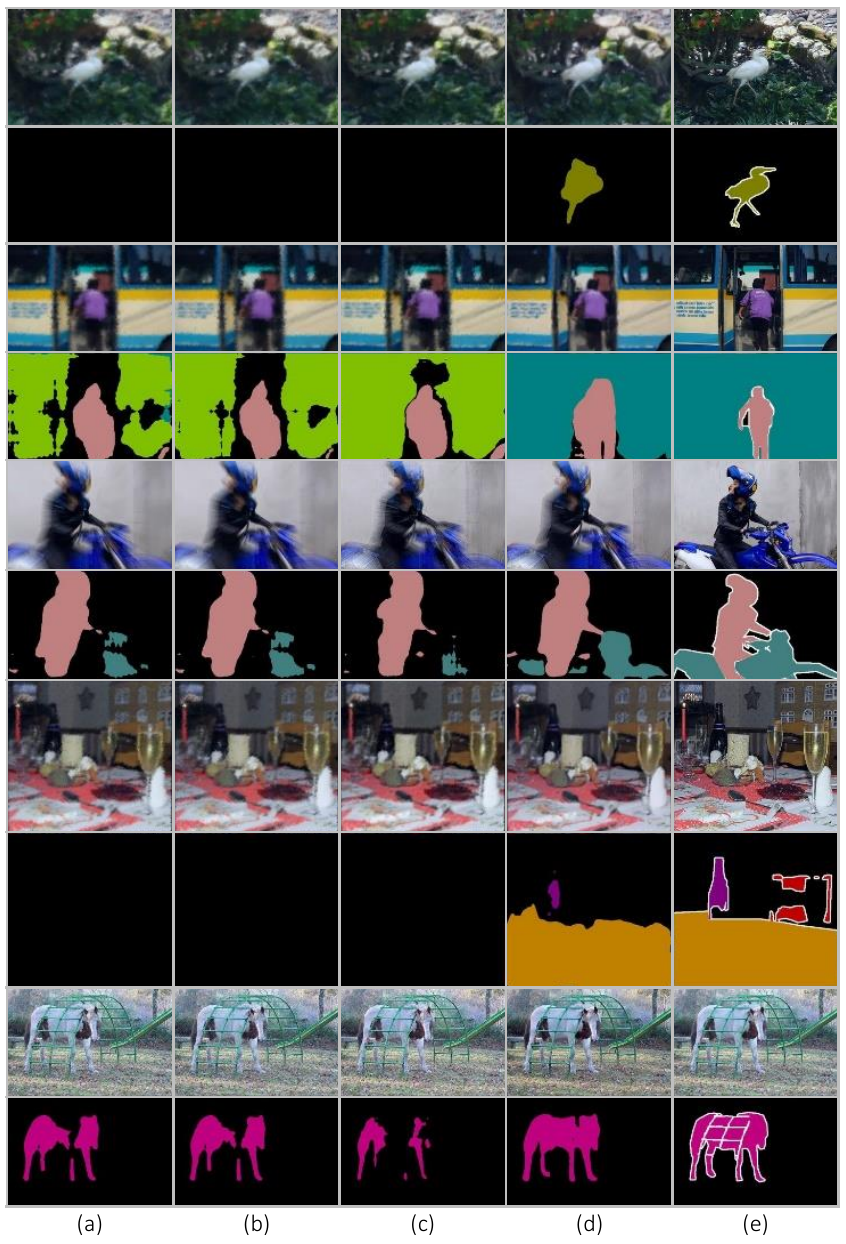}
\caption{
Additional qualitative results of DeepLab v3 on the VOC 2012 dataset. 
(a) Corrupted input. 
(b) OWAN~\cite{OWAN_CVPR19}. 
(c) URIE-MSE. 
(d) URIE. 
(e) Ground-truth. 
}
\vspace{-1mm}
\label{fig:seg_2}
\end{figure*}

\begin{figure*} [!t]
\centering
\includegraphics[width = 1 \textwidth] {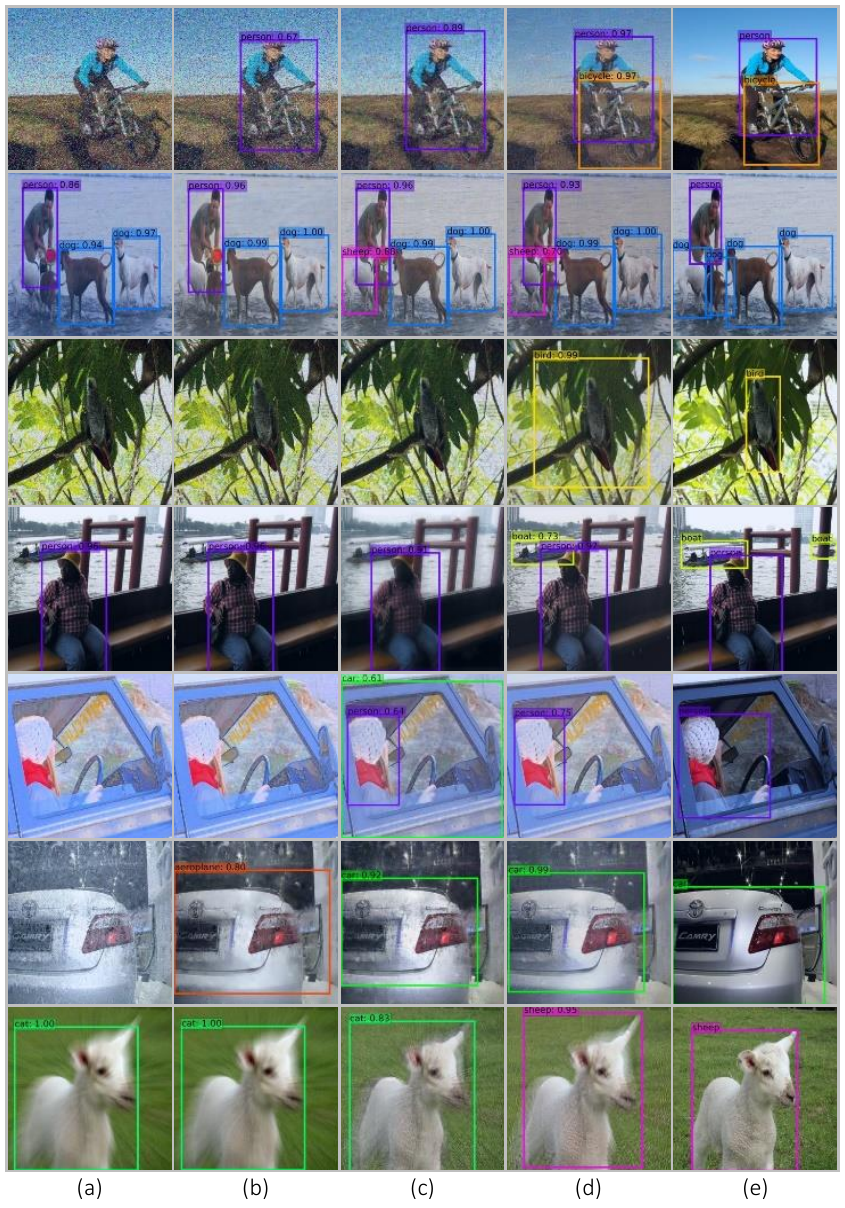}
\caption{
Additional qualitative results of SSD 300 on the VOC 2007 dataset. 
(a) Corrupted input.
(b) OWAN~\cite{OWAN_CVPR19}. 
(c) URIE-MSE. 
(d) URIE. 
(e) Ground-truth.
}
\vspace{-1mm}
\label{fig:det_1}
\end{figure*}

\begin{figure*} [!t]
\centering
\includegraphics[width = 1 \textwidth] {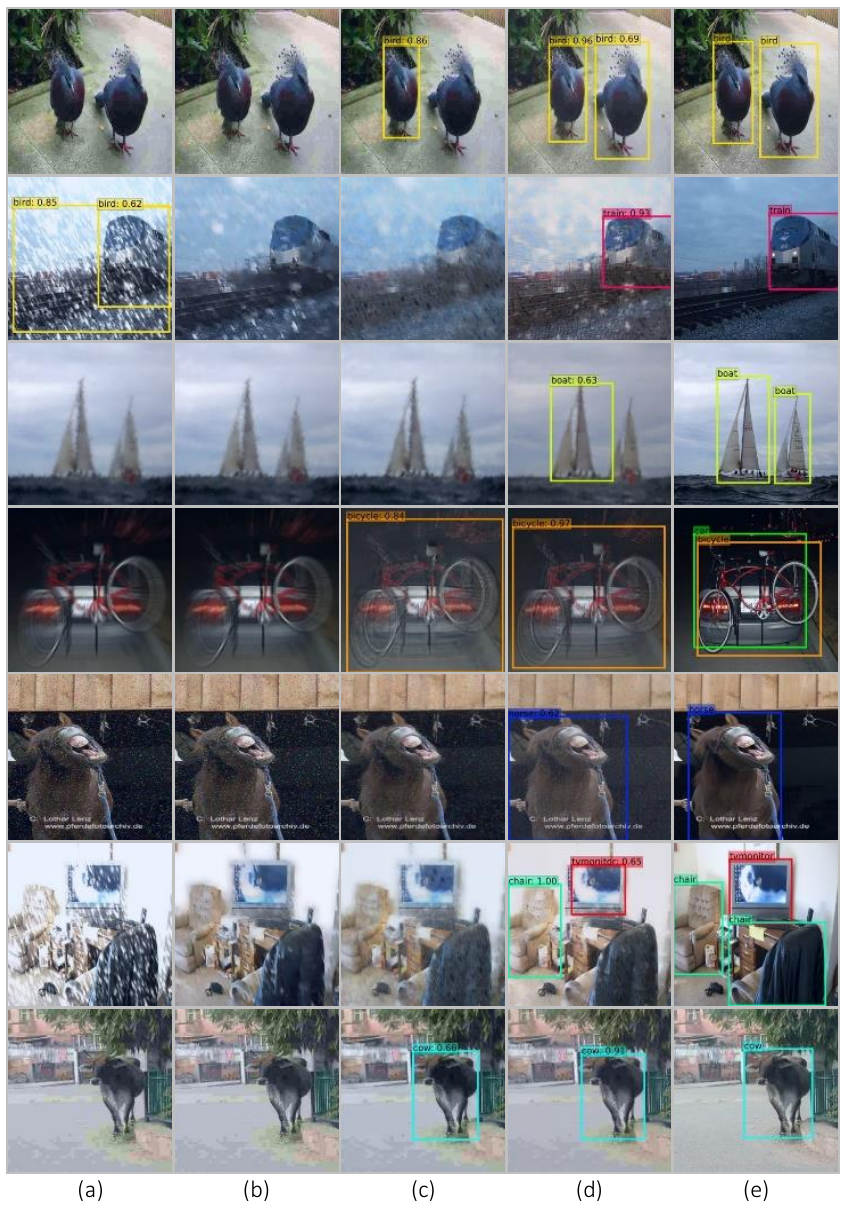}
\caption{
Additional qualitative results of SSD 300 on the VOC 2007 dataset. 
(a) Corrupted input.
(b) OWAN~\cite{OWAN_CVPR19}. 
(c) URIE-MSE. 
(d) URIE. 
(e) Ground-truth.
}
\vspace{-1mm}
\label{fig:det_2}
\end{figure*}

This section presents more qualitative results omitted in the main paper due to the space limit. 
\Fig{seg_1}~and~\Fig{seg_2} show the results of URIE on the PASCAL VOC 2012~\cite{Pascalvoc} semantic segmentation dataset.
They show that URIE can enhance images while focusing more on object-like areas instead of background. Also, the results not always looking plausible, especially when compared to those of the other methods, but directly improve the performance of the following semantic segmentation model.
\Fig{det_1}~and~\Fig{det_2} exhibit results on the PASCAL VOC 2007~\cite{Pascalvoc} object detection dataset.
Likewise, URIE best recovers salient regions of the images and improves the robustness of the object detector.


\begin{table}[!t] \small
\centering
\caption{
Classification accuracy on the ImageNet dataset. The numbers in parentheses indicate the differences from the performance of URIE. 
V16, R50, and R101 denote VGG-16, ResNet-50, and ResNet-101, respectively.
}
\scalebox{0.75}{
\begin{tabular}{@{}C{1cm}|@{}C{1.3cm}@{}C{1.3cm}@{}C{1.3cm}|@{}C{1.8cm}@{}C{1.8cm}@{}C{1.8cm}@{}|@{}C{1.8cm}@{}C{1.8cm}@{}C{1.8cm}@{}}
\multirow{2}{*}{}   & \multicolumn{3}{c|}{URIE}  & \multicolumn{3}{c|}{URIE-1/4} & \multicolumn{3}{c}{URIE-1/16} \\ 
\cline{2-10} & \multicolumn{1}{c}{Clean} & \multicolumn{1}{c}{Seen} & \multicolumn{1}{c|}{Unseen} & \multicolumn{1}{c}{Clean} & \multicolumn{1}{c}{Seen} & \multicolumn{1}{c|}{Unseen} & \multicolumn{1}{c}{Clean} & \multicolumn{1}{c}{Seen} & \multicolumn{1}{c}{Unseen}\\ \hline
V16 & 67.1 & 42.4 & 44.8 & 64.9 \red{(-2.2)} & 39.3 \red{(-5.0)} & 43.2 \red{(-1.6)} & 64.3 \red{(-2.8)} & 35.1 \red{(-7.3)} & 40.6 \red{(-4.2)}\\
R50 & 72.9 & 55.1 & 56.5 & 71.5 \red{(-1.4)} & 51.5 \red{(-2.5)} & 54.1 \red{(-1.0)} & 71.0 \red{(-1.9)} & 47.3 \red{(-7.8)} & 51.6 \red{(-4.9)}\\
R101& 74.1 & 57.8 & 59.4 & 72.8 \red{(-1.3)} & 54.9 \red{(-1.9)} & 57.2 \red{(-2.2)} & 72.4 \red{(-1.7)} & 51.2 \red{(-6.6)} & 55.8 \red{(-3.6)}
\end{tabular}
}
\label{tab:classificaion_on_imagenet_quarter}
\end{table}

\begin{table}[!t] \small
\centering
\caption{
Classification accuracy on the CUB dataset. The numbers in parentheses indicate the differences from the performance of URIE. V16, R50, and R101 denote VGG-16, ResNet-50, and ResNet-101, respectively.
}
\scalebox{0.75}{
\begin{tabular}{@{}C{1cm}|@{}C{1.3cm}@{}C{1.3cm}@{}C{1.3cm}|@{}C{1.8cm}@{}C{1.8cm}@{}C{1.8cm}@{}|@{}C{1.8cm}@{}C{1.8cm}@{}C{1.8cm}@{}}
\multirow{2}{*}{}   & \multicolumn{3}{c|}{URIE}  & \multicolumn{3}{c|}{URIE-1/4} & \multicolumn{3}{c}{URIE-1/16} \\ 
\cline{2-10} & \multicolumn{1}{c}{Clean} & \multicolumn{1}{c}{Seen} & \multicolumn{1}{c|}{Unseen} & \multicolumn{1}{c}{Clean} & \multicolumn{1}{c}{Seen} & \multicolumn{1}{c|}{Unseen} & \multicolumn{1}{c}{Clean} & \multicolumn{1}{c}{Seen} & \multicolumn{1}{c}{Unseen} \\ \hline
V16 & 77.9 & 58.8 & 48.9 & 76.8 \red{(-1.1)} & 57.1 \red{(-1.7)} & 48.0 \red{(-0.9)} & 76.8 \red{(-1.1)} & 53.7 \red{(-5.1)} & 45.8 \red{(-3.1)}  \\
R50 & 83.7 & 64.7 & 54.9 & 83.2 \red{(-0.5)} & 62.4 \red{(-2.3)} & 54.5 \red{(-0.4)} & 82.7 \red{(-1.0)} & 58.4 \red{(-6.3)} & 53.4 \red{(-1.5)} \\
R101& 84.1 & 67.1 & 56.6 & 83.2 \red{(-0.9)} & 64.8 \red{(-2.3)} & 56.3 \red{(-0.3)}  & 83.5 \red{(-0.6)} & 61.7 \red{(-5.4)} & 54.5 \red{(-2.1)} 
\end{tabular}
}
\label{tab:classificaion_on_cub_quarter}
\end{table}

\begin{table}[!t] \small
\centering
\caption{
Object detection performance of SSD 300 in mAP (\%) on the VOC 2007 dataset. 
The numbers in parentheses indicate the differences from the scores of URIE.
}
\scalebox{0.75}{
\begin{tabular}{@{}C{1.3cm}@{}C{1.3cm}@{}C{1.3cm}|@{}C{1.8cm}@{}C{1.8cm}@{}C{1.8cm}@{}|@{}C{1.8cm}@{}C{1.8cm}@{}C{1.8cm}@{}}
\multicolumn{3}{c|}{URIE-BN}  & \multicolumn{3}{c|}{URIE-1/4} & \multicolumn{3}{c}{URIE-1/16}  \\ \hline
\multicolumn{1}{c}{Clean} & \multicolumn{1}{c}{Seen} & \multicolumn{1}{c|}{Unseen} & \multicolumn{1}{c}{Clean} & \multicolumn{1}{c}{Seen} & \multicolumn{1}{c|}{Unseen} & \multicolumn{1}{c}{Clean} & \multicolumn{1}{c}{Seen} & \multicolumn{1}{c}{Unseen} \\ \hline
 76.5 & 59.4 & 62.7 & 75.3 \red{(-1.2)} & 56.8 \red{(-2.6)} & 61.1 \red{(-1.6)} & 75.2 \red{(-1.3)} & 53.0 \red{(-6.4)} & 59.4 \red{(-3.3)} \\
\end{tabular}
}
\label{tab:detection_on_voc_quarter}
\end{table}

\begin{table}[!t] \small
\centering
\caption{
Semantic segmentation performance of DeepLab v3 in mIoU (\%) on the VOC 2012 dataset.
The numbers in parentheses indicate the differences from the score of URIE.
}
\scalebox{0.75}{
\begin{tabular}{@{}C{1.3cm}@{}C{1.3cm}@{}C{1.3cm}|@{}C{1.8cm}@{}C{1.8cm}@{}C{1.8cm}@{}|@{}C{1.8cm}@{}C{1.8cm}@{}C{1.8cm}@{}}
\multicolumn{3}{c|}{URIE}  & \multicolumn{3}{c|}{URIE-1/4} & \multicolumn{3}{c}{URIE-1/16}  \\ \hline
\multicolumn{1}{c}{Clean} & \multicolumn{1}{c}{Seen} & \multicolumn{1}{c|}{Unseen} & \multicolumn{1}{c}{Clean} & \multicolumn{1}{c}{Seen} & \multicolumn{1}{c|}{Unseen} & \multicolumn{1}{c}{Clean} & \multicolumn{1}{c}{Seen} & \multicolumn{1}{c}{Unseen} \\ \hline
 78.6 & 67.0 & 67.2 & 78.5 \red{(-0.1)} & 65.1 \red{(-1.9)} & 67.3 \green{(+0.1)} & 78.4 \red{(-0.2)} & 62.7 \red{(-4.3)} & 65.3 \red{(-1.9)}  \\
\end{tabular}
}
\label{tab:segmentation_on_voc_quarter}
\end{table}


\begin{table}[!t]
\centering
\caption{
Restoration performance on the CUB dataset in terms of MSE and SSIM. Their accuracies over CUB dataset are presented alongside. 
}
\begin{tabular}{@{}C{2.1cm}|@{}C{1.3cm}@{}C{1.3cm}@{}C{1.7cm}}
          & MSE   & SSIM  & Recog. Acc. \\ \hline
URIE      & 0.331 & 0.358 & 55.1        \\
URIE-MSE  & 0.158 & 0.445 & 44.5        \\
URIE-SSIM & 0.206 & 0.445 & 44.6        \\
OWAN~\cite{OWAN_CVPR19}      & 0.380 & 0.366 & 42.6       
\end{tabular}
\label{tab:restoration_performance}
\end{table}

\clearpage

\end{document}